\documentclass[acmtog,screen,authorversion]{acmart} 
\graphicspath{{./images/}}
\usepackage{color,hyphenat,balance,booktabs}
\usepackage{graphicx}
\usepackage{xcolor}
\usepackage{overpic}
\usepackage{autonum}
\usepackage{wrapfig}
\usepackage{physics}
\usepackage{amsmath}
\usepackage[ruled,linesnumbered]{algorithm2e}

\SetCommentSty{mycommfont}
\usepackage{enumitem}
\setlist{nosep}
\newtheorem*{definition*}{Definition}
\theoremstyle{remark}
\newtheorem*{remark}{Remark}

\def \etal {{\emph{et al}.\thinspace}}
\def \cf {{\emph{cf}.\thinspace}}
\def \etc {{\emph{etc}.\thinspace}}
\def \eg {{\emph{e.g}.\thinspace}, }
\def \ie {{\emph{i.e}.\thinspace}, }

\setcitestyle{square}

\acmSubmissionID{papers\_222}

\setcopyright{acmlicensed}\acmJournal{TOG}
\acmYear{2022}\acmVolume{41}\acmNumber{4}\acmArticle{129}\acmMonth{7}
\acmDOI{10.1145/3528223.3530078}

\newcommand{\pluseq}{\mathrel{+}=}
\newcommand{\name}{ComplexGen\xspace}
\newcommand{\netname}{ComplexNet\xspace}

\begin{document}

\title{ComplexGen: CAD Reconstruction by B-Rep Chain Complex Generation}
\author{Haoxiang Guo}
\affiliation{
	\institution{Tsinghua University}
	\country{}
}
\affiliation{
    \institution{Microsoft Research Asia}
    \country{China}
}
\email{ghx17@mails.tsinghua.edu.cn}
\author{Shilin Liu}
\affiliation{
	\institution{University of Science and Technology of China}
	\country{}
}
\affiliation{
    \institution{Microsoft Research Asia}
    \country{China}
}
\email{freelin@mail.ustc.edu.cn}
\author{Hao Pan}
\email{haopan@microsoft.com}
\author{Yang Liu}
\email{yangliu@microsoft.com}
\author{Xin Tong}
\email{xtong@microsoft.com}
\author{Baining Guo}
\email{baining@microsoft.com}
\affiliation{
	\institution{Microsoft Research Asia}
	\country{China}
}
\authorsaddresses{Authors' addresses: H.~Guo, Tsinghua University, Haidian District, Beijing, China; ghx17@mails.tsinghua.edu.cn; S.~Liu, University of Science and Technology of China, Hefei, China; freelin@mail.ustc.edu.cn; H.~Pan (corresponding author), Y.~Liu, X.~Tong, B.~Guo, 5 Danling St., Haidian District, Beijing, China; \{haopan, yangliu, xtong, baining\}@microsoft.com. }

\begin{teaserfigure}
	\centering
	\begin{overpic}[width=0.9\linewidth]{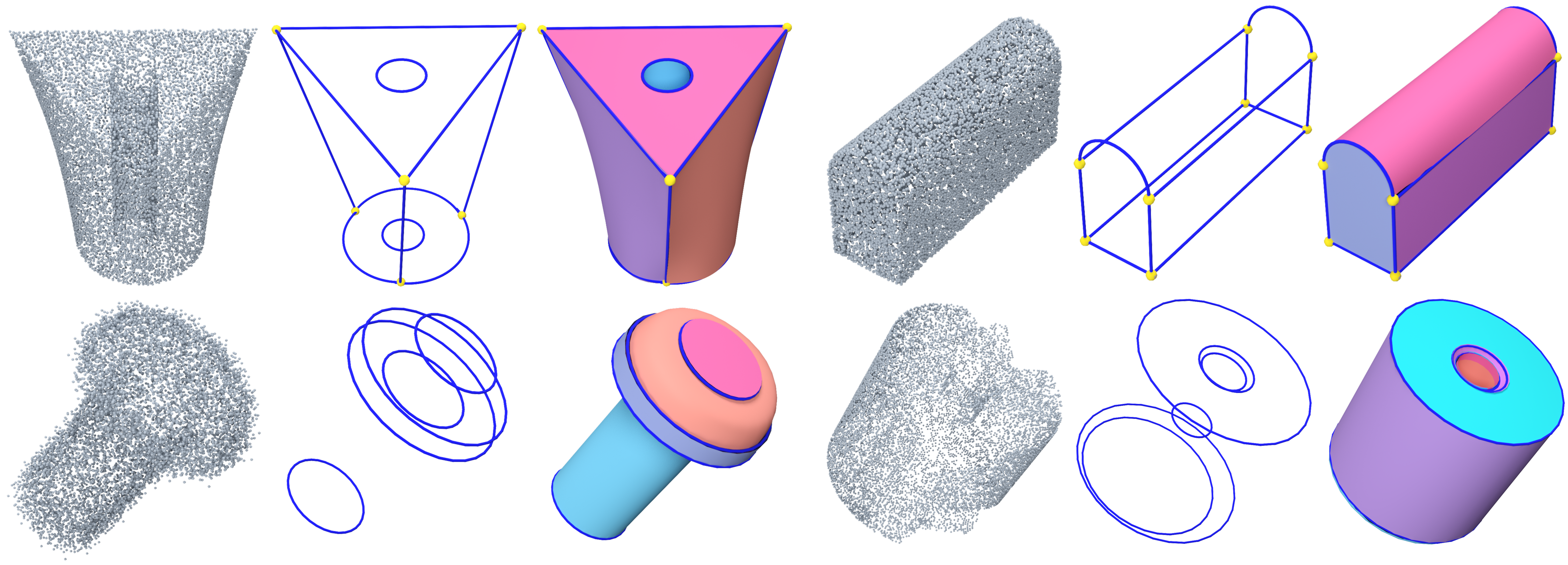}
	\put(0, 21){\small (a)}
	\put(50, 21){\small (b)}
	\put(0, 1){\small (c)}
	\put(50, 1){\small (d)}
	\end{overpic}
	\caption{\textbf{\name for CAD reconstruction from point clouds}. 
	Given an input point cloud, \name recovers corners, curves and patches simultaneously along with their mutual topology constraints, which enables more complete, regularized and structured CAD reconstruction in the boundary representation (B-Rep).
	For each example, the input points, the reconstructed corners (yellow) and curves (blue) and the full B-Rep models (surface patched randomly colored) are shown. The input points for (c) are corrupted by noise and those for (d) are only partial. 
	}
	\label{fig:teaser}
\end{teaserfigure}
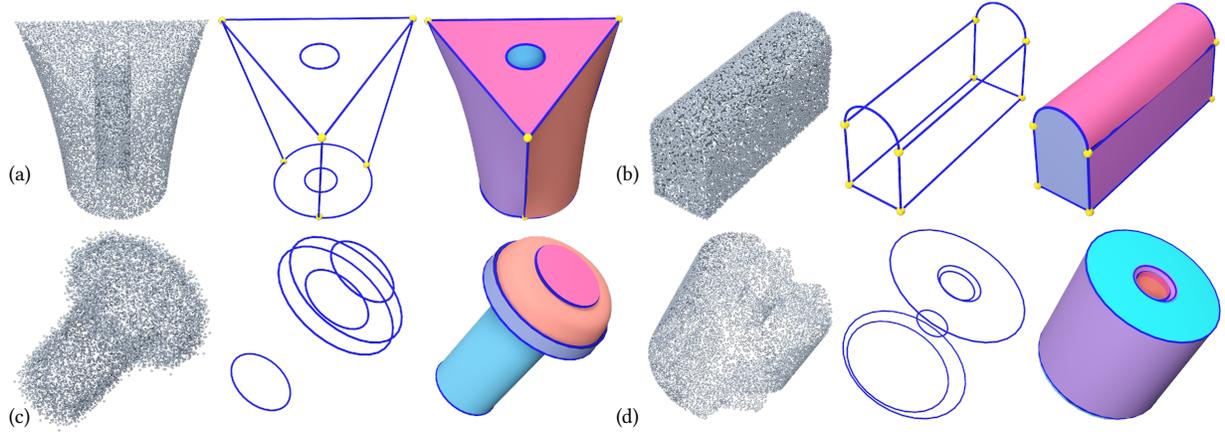

\begin{abstract}

We view the reconstruction of CAD models in the boundary representation (B-Rep) as the detection of geometric primitives of different orders, \ie vertices, edges and surface patches, and the correspondence of primitives, which are holistically modeled as a chain complex, and show that by modeling such comprehensive structures more complete and regularized reconstructions can be achieved.
We solve the complex generation problem in two steps.
First, we propose a novel neural framework that consists of a sparse CNN encoder for input point cloud processing and a tri-path transformer decoder for generating geometric primitives and their mutual relationships with estimated probabilities.
Second, given the probabilistic structure predicted by the neural network, we recover a definite B-Rep chain complex by solving a global optimization maximizing the likelihood under structural validness constraints and applying geometric refinements.
Extensive tests on large scale CAD datasets demonstrate that the modeling of B-Rep chain complex structure enables more accurate detection for learning and more constrained reconstruction for optimization, leading to structurally more faithful and complete CAD B-Rep models than previous results.

\end{abstract}




\maketitle

\section{Introduction}

Reconstructing a concise and structured CAD representation from an unstructured point cloud is a fundamental problem, commonly known as \textit{reverse engineering} with a long history of extensive research and commercial applications \cite{ansys,3dsystem}.
To this end, previous works typically try to segment the input point cloud into regions and fit them with geometric primitives of different types and parameters.
For example, recent learning-based methods segment input points into non-overlapping regions and fit them with typed primitive surface patches \cite{LiPrimitive2019CVPR,ParseNet2020ECCV,yan2021hpnet}, or detect sharp corner and curves from which a wireframe is connected \cite{Wang2020pienet,Liu2021pcwf}.
Despite the significantly improved reconstruction accuracy over heuristic approaches for respective geometric primitives, the simultaneous reconstruction of patches, curves and corners is not addressed by these works, leading to incomplete reconstructions.
Based on such incomplete reconstructions, it is nontrivial to establish a consistent structure with conforming elements and proper topology that is desirable for CAD models.

In this paper, we present the \name framework that generates complete structures consisting of primitives and their explicit topology relationships comprehensively.
The modeling of complete structures is not only a desirable target by itself, but also enables more robust CAD reconstruction.
Specifically in our framework, the explicit generation of topology and geometry effectively divides the task into two sub-problems: the detection of elements and their topology, as well as the regression of their geometric details.
The division brings several benefits.
First, the sub-tasks are more amenable to deep learning as standard categorical or regression tasks, where the topology reconstruction is categorical and the geometry embedding is regression.
Second, the combination of topology and geometry provides complementary information for more constrained CAD reconstruction.
In particular, the explicit modeling of topology allows us to introduce constraints that ensure topological validity and result in more complete reconstructions.

To model such a comprehensive structure, we take the boundary representation (B-Rep) as the basic structure, which is fundamental for CAD applications \cite{hoffmann1989geometric}.
The B-Rep structure consists of elements of different orders, \ie vertices, edges and faces, and the topology among the elements.
We show that the B-Rep defines a chain complex \cite{Hatcher:AT} that reveals structural constraints, \eg the boundary curves of a face form closed loops, which must be satisfied for structurally valid CAD model reconstruction.

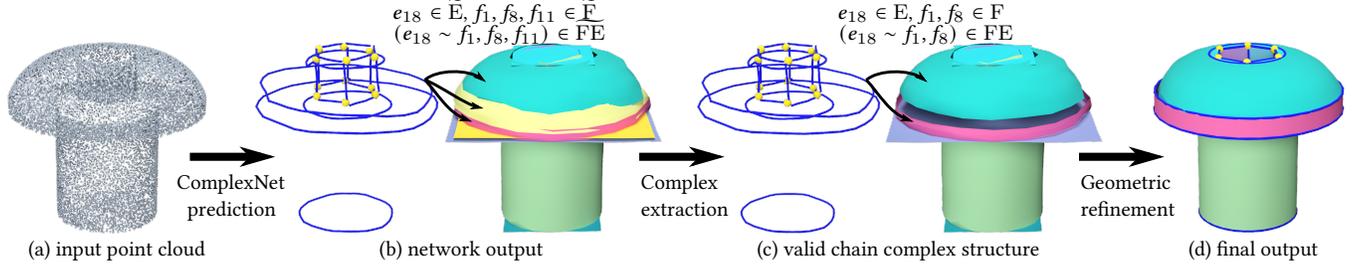
\begin{figure*}
	\vspace{1mm}
	\begin{overpic}[width=\linewidth]{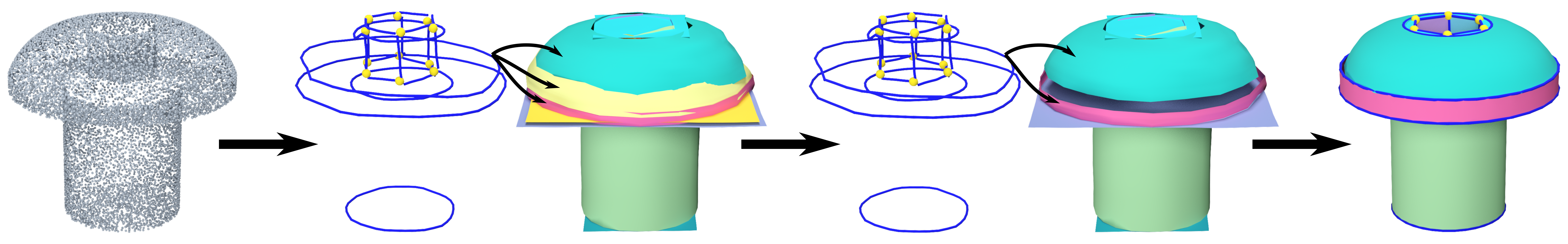}
	\put(2,-1){\small (a) input point cloud}
	\put(28,-1){\small (b) network output}
	\put(56,-1){\small (c) valid chain complex structure}
	\put(88,-1){\small (d) final output}
	\put(13,4){\small \netname}
	\put(13.8,2){\small prediction}
	\put(47.4,4){\small Complex}
	\put(47.4,2){\small extraction}
	\put(80,4){\small Geometric}
	\put(80,2){\small refinement}
	%\put(29,19){\small $e_{18}\in \widetilde{\mathbf{E}}$}
	\put(29,16.6){\small $e_{18}\in \widetilde{\mathbf{E}}$, $f_1,f_8,f_{11}\in \widetilde{\mathbf{F}}$}
	\put(29,15){\small $\left(e_{18} \sim f_1,f_8,f_{11}\right) \in \widetilde{\mathbf{FE}}$}
	%\put(62,19){\small $e_{18}\in \mathbf{E}$}
	\put(62,16.6){\small $e_{18}\in \mathbf{E}$, $f_1,f_8\in \mathbf{F}$}
	\put(62,15){\small $\left(e_{18} \sim f_1,f_8\right)\in \mathbf{FE}$}
	\end{overpic}
	\caption{\textbf{Pipeline of \name}. The point cloud (a) first goes through the \netname with sparse CNN encoder and transformer decoders to generate primitive elements of different orders, i.e. corners, curves and faces, and their mutual topology.
	(b) shown here are the elements with validness probabilities above 0.5; the corners/curves and patches are shifted apart for better visibility. The curve $e_{18}$ and its adjacent patches are highlighted as an example of the predicted topology. 
	(c) the probabilistic B-Rep chain complex then goes through a global optimization that solves for the optimal connection and existence of elements, where the patches adjacent to $e_{18}$ are resolved by removing the redundant $f_{11}$.
	(d) the structure is refined geometrically by fitting each element shape to input points, mutual constraints and given types; the refined elements collectively constitute a structurally valid B-Rep model. }
	\label{fig:pipeline}
\end{figure*}

To generate a B-Rep chain complex from a point cloud requires the detection of geometric primitives of different orders and the reasoning of their topology relationships, which we address with a hybrid approach (Fig.~\ref{fig:pipeline}).
First, we design \netname, a neural network for learning to generate B-Rep structures from point clouds. 
In \netname, we use a sparse CNN to extract spatial features from the input point cloud and three paths of transformer decoder networks to output the three groups of primitive elements, including their shapes and geometric types (circles, planes, \etc), and their mutual topology defined as adjacency matrices.
We connect the three element generation paths throughout the decoder layers, to enable mutual communication and facilitate the emergence of a consistent B-Rep model.
Second, given the network predicted B-Rep structure, which is imprecise and probabilistic in nature, we extract a precise and valid B-Rep model through global optimization that involves combinatorial valid topology extraction and constrained geometric fitting. 

The modeling of holistic B-Rep chain complex structures enables more accurate detection and topology reconstruction in the first stage of network prediction (Sec.~\ref{sec:ablation}), and allows to formulate better constrained global optimization problems where diverse conditions like the face boundary loop closedness, edge manifoldness, cylinder/circle perpendicularity, \etc can be applied for more regular and complete reconstruction (Sec.~\ref{sec:postproc}).
These enhancements demonstrate the benefits that structures incorporated by \name can bring to CAD reconstruction.

We test our approach on a large scale benchmark from the ABC dataset \cite{Koch2019ABC}.
Compared with previous methods that only attack parts of the B-Rep reconstruction, our method recovers the total B-Rep chain complex structures consistently, with part accuracy comparable to or superior than previous methods.
Challenging tests with noisy and incomplete point clouds also demonstrate that our joint reasoning of different orders of elements improves robustness to data corruption.

To summarize, we make the following contributions in this paper:
\begin{itemize}[leftmargin=10pt,itemsep=1pt]
    \item CAD reconstruction by B-Rep chain complex generation. We define a more structured reconstruction task where elements of different orders and their topology relationships are considered simultaneously, and show its benefits in recovering more complete reconstructions.
    \item A deep neural network for generating the B-Rep chain complex models. We generate the three groups of elements and their topology with a tri-path structure that promotes co-occurrence of consistent elements. 
    \item Neurally guided global reconstruction with enhanced structural constraints.
    Given the neural network predicted probabilistic structure, we solve a tractable global optimization problem to recover definite CAD B-Rep models with structure validness. 
\end{itemize}
We open-source the code and data to facilitate future research\footnote{The repository is available at https://github.com/guohaoxiang/ComplexGen.}.

\section{Related work}

\paragraph{CAD reconstruction}

Primitive based CAD reconstruction has critical industrial applications and therefore an extensive literature in computer graphics \cite{SRSurvey2017} and reverse engineering \cite{Werghi02}.
While the problem inherently involves both combinatorial search of geometric primitives and continuous optimization of data fitting, we roughly categorize the existing approaches into the traditional methods that focus more on numerical  optimization and the more recent methods that use neural networks learning data driven priors to overcome the overwhelming complexity, as discussed below.

To fit candidate primitive surface patches to the input shape discretized as a point cloud or polygonal mesh, early works apply probabilistic search like RANSAC \cite{Schnabel2007,Schnabel2009} or solve variational optimizations \cite{VSA2004,VSAPS2020,Yan2012} that iterate between primitive fitting and segmentation. 
Subsequent works apply global optimization over structural constraints to enhance the local primitive fitting as well as the overall model structural regularity \cite{GlobFit2011,NanPolyFit2017}.
However, the overwhelming combinatorial complexity of the CAD reconstruction problem makes such approaches prone to locally optimal solutions.  

With the advent of versatile deep learning methods, renewed efforts have been made to address the CAD reconstruction problem with data-driven methods.
In particular, Li \etal~\shortcite{LiPrimitive2019CVPR}, Sharma \etal~\shortcite{ParseNet2020ECCV}, Yan \etal~\shortcite{yan2021hpnet} and Huang \etal~\shortcite{Huang_2021_ICCV} train point-based neural networks to assign patch primitive types and parameters to each input point, which effectively solves the segmentation and fitting problem in one pass, by learning on large-scale annotated CAD data sets.
Liu \etal~\shortcite{Liu2021pcwf} and Wang \etal~\shortcite{Wang2020pienet} instead reconstruct the wireframes that delineate the different primitive patches, by first detecting the candidate corners and lines from input points, and then connecting them into coherent wire frames through exhaustive enumeration and checking.
Nash \etal~\shortcite{Nash2020polygen} focus on polygonal model reconstruction, and uses an auto-regressive network to predict the vertices and planar faces iteratively.

We note that the separation in terms of segmentation and fitting, as well as the separation of corner, edge and patch reconstruction into different stages, are sub-optimal by preventing mutual data exchange.
Instead, we propose a holistic B-Rep chain complex representation that entails the elements of multiple orders as well as their topology, and recover such a representation by a unified neural framework.
In addition, our global optimization guided by network prediction further improves structural regularity and validness with constraints on topology and geometric fitting, which is missing from other learning-based methods as they focus on only parts of the whole B-Rep model and cannot formulate such constraints.

A different approach to CAD reconstruction is to solve the inverse problems of procedural CAD models.
For example, inverse CSG \cite{InverseCSG2018,CSGNet2018,UCSGNet2020,CSGStump2021} searches for CSG boolean operations and solid primitives that combine into the target shape; \cite{DeepCAD2021,WillisSketch2021,seff2021vitruvion,CADasLang2021,SketchGen2021} assume a ``sketch+extrude'' procedural model and study the generation of 2D sketches that can be extruded to obtain 3D shapes, by training on datasets of such modeling sequences \cite{willis2020fusion,SketchGraphs}. 
While expressive within their scopes, these procedural models do not cover all CAD operations and have difficulty modeling e.g. freeform curves and surfaces resulting from blending, beveling, revolution and other operations (\cf Figs.~\ref{fig:teaser}(a), \ref{fig:result_gallery}(a), \ref{fig:comparison_PIENET}(b), \ref{fig:noise_partial_data}(a), \etc).
Different from procedural models, B-Rep is a widely used underlying representation for CAD models.
Indeed, with reconstructed B-Rep models, more advanced reverse engineering tasks like command sequence reconstruction and editing can be enabled \cite{ZoneGraph2021,lambourne2021brepnet,jayaraman2021uvnet,Willis2021JoinABLeLB,ConstraintSynthesis2021,cascaval2021differentiable}.

\paragraph{Transformer based detection}

The object detection and segmentation pipeline has recently observed a major shift, from the previous CNN-based methods that make object proposals relative to anchor points in the input \cite{FasterRCNN2017,MaskRCNN2017,Yolo2016}, to the emerging methods that output absolute positions of detected objects directly by conditioning on the input \cite{DETR2020}, which is also shown effective in 3D \cite{Misra3DETR2021}.
Indeed, the above mentioned learning-based CAD reconstruction methods \cite{LiPrimitive2019CVPR,ParseNet2020ECCV,Wang2020pienet,Liu2021pcwf,yan2021hpnet,Huang_2021_ICCV} reflect the traditional scheme of CNN-based detection, by first classifying the input points as potential anchors of segments and then merging them into groups through clustering and searching.
In contrast, PolyGen \cite{Nash2020polygen} employs the transformer backbone to auto-regressively output the absolute vertex positions and their grouping into faces, which is less restricted by the input point accuracy.
We build on the DETR framework \cite{DETR2020} to detect each class of elements directly.
However, a naive adaption of the framework does not show consistency across the elements or robust geometric accuracy (Sec.~\ref{sec:ablation}).
Instead, we model the rich structures of a B-Rep chain complex by additionally generating its topology, which is shown to improve the detection accuracy and provide accurate topological structure of a B-Rep model.
Moreover, with our improved network the input point cloud can be heavily noisy and corrupted, which is common in CAD reconstruction from scanned data, while our prediction still leverages the overall structure and remains stable.
Finally, our chain complex structure prediction enables neurally guided global optimization that finalizes a manifold and complete B-Rep model.

\section{CAD models as B-Rep chain complexes}
\label{sec:brep_formulation}

CAD models are frequently represented in boundary representation (B-Rep) \cite{weiler1986topological,hoffmann1989geometric}, which defines the boundary of a solid 3D shape by elements of different orders, like faces, edges and vertices, that are linked together and geometrically embedded into the 3D space by primitive types for surfaces (\eg plane, sphere, cylinder) and curves (\eg line, circle, spline).
The separate and explicit representation of topology and geometry by B-Rep is originally designed to facilitate modeling and overcome geometric inaccuracies \cite{weiler1986topological}.
In this section, we show that B-Rep naturally admits a chain complex structure with computable validity constraints.
We present the B-Rep chain complex formulation and computational implementation, and give an overview of its generation from a point cloud.

\subsection{B-Rep Chain Complex}

We denote a B-Rep model of order 3 as $\mathcal{C} = (V,E,F, \partial, \mathcal{P})$ with 0th-order vertices $V=\{v_i\}$, 1st-order curved edges $E=\{e_i\}$ that may or may not be closed, and 2nd-order curved faces $F=\{f_i\}$; 
the elements of different orders are connected by the graded boundary operator $\partial_{n}$, $n=1,2$.
Therefore, $\partial_2 f_i \subset E$ gives the edges which define the boundary of face $f_i$, and $\partial_1 e_i \subset V$ gives the endpoint vertices of the edge $e_i$.

Each element set generates a vector space, i.e., $\mathbb{F} = \{\sum_{i} {\lambda_if_i} | \lambda_i \in \mathbb{Z}, f_i\in F\}$ and similarly for $\mathbb{E}, \mathbb{V}$. 
When the elements are oriented, the boundary operators define linear maps between the spaces, \ie $\partial_2: \mathbb{F} \rightarrow \mathbb{E}$ with basis transform $\partial_2 f_i = \sum_{j}{\lambda_{i,j} e_j}$, where $\lambda_{i,j}\in\{0,\pm 1\}$ denotes whether the orientation of the face $f_i$ induces positive or negative orientation of its boundary edge $e_{i,j}$, or 0 for a non-adjacent edge.
Similar definition applies to $\partial_1: \mathbb{E} \rightarrow \mathbb{V}$.
The graded vector spaces of elements and boundary operators form a chain complex \cite{Hatcher:AT}:
\begin{equation}
	\mathbb{F} \xrightarrow{\partial_2} \mathbb{E} \xrightarrow{\partial_1} \mathbb{V}
\end{equation}
with the property $\partial_1\circ\partial_2 = 0$.
Intuitively the property establishes that for each B-Rep face, its associated edges form closed loops by joining heads with tails. 
A similar treatment of B-Rep as chain complexes is discussed in \cite{AlgebraicRep2014}; however, our formulation of constraints (Eqs.~(\ref{eq:opt:cons:manifold})-(\ref{eq:opt:cons:closed})) for reconstruction is novel. 

%in the sense that each of the $V,E,F$ set define an Abelian group with addition operation.

The above topological structure is further augmented with geometric data $\mathcal{P}$ that associates primitive shapes to each element.
To accommodate available datasets, we restrict to the following types of surfaces: \textit{plane, cylinder, torus, B-spline, cone, sphere}, and the following types of curves: \textit{line, circle, B-spline, ellipse}.
For a primitive type, there are parameters defining its geometry, as well as properties such as if an edge curve is \textit{open} or \textit{closed}, and if a face patch is \textit{open} or \textit{closed}.
The set of geometric primitives considered here is inline with \cite{ParseNet2020ECCV} and larger than many other works \cite{Liu2021pcwf,Wang2020pienet,LiPrimitive2019CVPR}. 

Throughout the discussions, we use vertex/edge/face and corner/curve/patch interchangeably for naming the three orders of elements, with the former primarily used for denoting the topological aspect and the latter for the geometrical aspect of a B-Rep chain complex. 
Frequent notations used are summarized in Table~\ref{tab:notations}.

\subsection{Structure Representation and Validness}

To encode the elements of a B-Rep chain complex, we use binary vectors of dimensions $N_v, N_e, N_f$ for vertices, edges and faces, respectively, \ie the binary vectors $\mathbf{F}\in\mathbb{B}^{N_f}, \mathbf{E}\in\mathbb{B}^{N_e}, \mathbf{V}\in\mathbb{B}^{N_v}$ denote if any member of a element group exists for a given input shape. 
The binary matrices $\mathbf{FE}\in\mathbb{B}^{N_f\times N_e}, \mathbf{EV}\in\mathbb{B}^{N_e\times N_v}, \mathbf{FV}\in\mathbb{B}^{N_f\times N_v}$ denote if the corresponding elements are adjacent, thus encoding the boundary operators $\partial_2,\partial_1$ and their composite $\partial_1\circ \partial_2$;
for non-existent elements, the corresponding rows and columns in the adjacency matrices are set zero. 
Furthermore, the binary vector $\mathbf{O}\in \mathbb{B}^{N_e}$ indicates if an edge is open-ended.
Note that by using binary matrices, we have discarded the orientation information of the elements.
This is to remove the orientation ambiguity for input point cloud without normal vectors (Sec.~\ref{sec:detection}) and to simplify global optimization by dealing with binary variables only (Sec.~\ref{sec:postproc}).

For a valid B-Rep chain complex that represents a closed manifold CAD model, we naturally arrive at the following properties:
\begin{align}
	\textstyle{\sum}_{i}{\mathbf{FE}[i,j]} &= 2\mathbf{E}[j], \quad j\in [N_e] \label{eq:opt:cons:manifold}\\
	\textstyle{\sum}_{j}{\mathbf{EV}[i,j]} &= 2\mathbf{E}[i]\mathbf{O}[i], \quad i\in [N_e] \label{eq:opt:cons:curveend}\\
	\mathbf{FE}\times \mathbf{EV} &= 2\mathbf{FV} \label{eq:opt:cons:closed}
\end{align}
The three equations can be described as follows:
\begin{enumerate}
	\item[(\ref{eq:opt:cons:manifold})]- an edge is adjacent to two faces;
	\item[(\ref{eq:opt:cons:curveend})]- an open/closed edge has two/zero endpoints;
	\item[(\ref{eq:opt:cons:closed})]- a face has closed boundary loops.
\end{enumerate}
(\ref{eq:opt:cons:closed}) is due to the B-Rep chain complex structure satisfying $\partial_1\circ\partial_2 = 0$, without assuming orientations of the elements.
Among these equations, both (\ref{eq:opt:cons:curveend}) and (\ref{eq:opt:cons:closed}) are quadratic and require special effort to implement with computational feasibility (\cf Sec.~\ref{sec:postproc}). 

On the other hand, for a structure satisfying the above three properties, \ie the edge manifoldness and the face boundary closedness, by induction on edges and faces, we have an algebraic B-Rep structure representing an edge-manifold surface.
Assuming successful geometric realization of such a topological structure, we arrive at the following intuitive definition of structure validness for a B-Rep chain complex.
\begin{definition*}
A B-Rep chain complex $\mathcal{C}$ is valid if and only if its topology satisfies the constraints (\ref{eq:opt:cons:manifold})-(\ref{eq:opt:cons:closed}) and its geometry realizes the topology with sufficient accuracy.
\end{definition*}
We propose a hybrid approach that combines deep learning and neurally guided optimization to solve the problem of valid B-Rep chain complex reconstruction from point clouds. 

\begin{table}
	\caption{Summary of notations frequently used in the paper.}\vspace{-6mm}
	\begin{tabular}{r l}
		\\ \hline 
		Notation & Description \\ \hline
		$\mathbb{B}$ & Boolean domain $\{0,1\}$\\
		$N_{v/e/f}$ & element numbers for vertices/edges/faces\\
		$d$ & the latent dimension for tri-path decoder\\
		$[N]$ & the sequence of integers from 1 to $N$ \\
		$\mathcal{C} = (V, E, F, \partial, \mathcal{P})$ & a B-Rep chain complex with vertices $V$, \\
		& \ \   edges $E$,  faces $F$, boundary operator $\partial$ \\
		& \ \   and geometry embedding $\mathcal{P}$ \\
		$\mathbf{V},\mathbf{E},\mathbf{F} \in \mathbb{B}^{N_{v/e/f}}$ & validness of generated elements\\
		$\mathbf{O} \in \mathbb{B}^{N_e}$ & openness of edges\\
		$\mathbf{FE}\in\mathbb{B}^{N_f\times N_e}$ & the adjacency of faces and edges\\
		$\mathbf{EV}\in\mathbb{B}^{N_e\times N_v}$ & the adjacency of edges and vertices\\
		$\mathbf{FV}\in\mathbb{B}^{N_f\times N_v}$ & the adjacency of faces and vertices\\
		$[\mathbf{S}_p {\in} \mathbb{R}^{S\times 3}, \mathbf{S}_f {\in} \mathbb{R}^{S\times d}]$ & CNN output voxels and features \\
		$\mathbf{Q}_{v/e/f} \in \mathbb{R}^{N_{v/e/f}\times d}$ & initial embeddings for generation \\
		$\mathbf{G}_{v/e/f} \in \mathbb{R}^{d}$ & class embeddings for element groups \\
		$M_{v/e/f}, m(\cdot), m'(\cdot)$ & generated elements matched with GT\\
		 & \ \   and forward/backward correspondences \\
		\hline
	\end{tabular}
	\label{tab:notations}
\end{table}

\begin{figure*}
	\begin{overpic}[width=\linewidth]{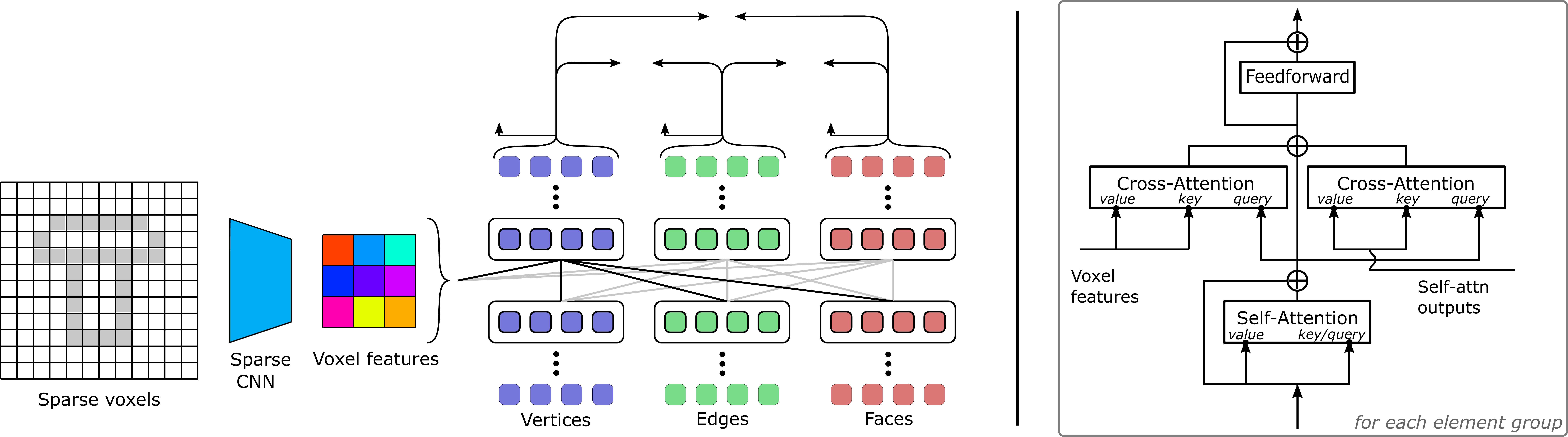}
		\put(73,8.5){\small $\mathbf{S}$}
		\put(69.5,12.5){\footnotesize $\mathbf{S}_f$}
		\put(72,12.5){\footnotesize $\mathbf{S}_f {+} \mathbf{S}'_p$}
		\put(78.3,12.5){\footnotesize LN,+$\mathbf{Q}_t$}
		\put(87,12.8){\footnotesize ${+}\mathbf{Q}_{t'}{+}G_{t'}$}
		\put(95,12.5){\footnotesize LN,+$\mathbf{Q}_t$}
		\put(77.5,4){\footnotesize LN}
		\put(86.5,4){\footnotesize LN,+$\mathbf{Q}_t$}
		
		\put(95,7.6){\small $\mathbf{X}_{t'\neq t}$}
		\put(80.6,0.7){\small $\mathbf{H}^i_{t}$}
		\put(79.5,26.3){\small $\mathbf{H}^{i+1}_{t}$}
		
		\put(83,20.4){\footnotesize LN}
		
		\put(61.5,6.9){\small $\mathbf{H}^i_{t}$}
		\put(61.5,12.1){\small $\mathbf{H}^{i+1}_{t}$}
		\put(61.5,2.3){\small $\mathbf{Q}_{t}$}
		\put(30.5,20.5){\small $\widetilde{\mathbf{V}}, \widetilde{\mathbf{P}_v}$}
		\put(38,20.5){\small $\widetilde{\mathbf{E}}, \widetilde{\mathbf{P}_e}, \widetilde{\mathbf{T}_e}, \widetilde{\mathbf{O}}$}
		\put(49,20.5){\small $\widetilde{\mathbf{F}}, \widetilde{\mathbf{P}_f}, \widetilde{\mathbf{T}_f}, \widetilde{\mathbf{U}}$}
		\put(45.1,26.5){\small $\widetilde{\mathbf{FV}}$}
		\put(39.6,23.5){\small $\widetilde{\mathbf{EV}}$}
		\put(50.5,23.5){\small $\widetilde{\mathbf{FE}}$}
		
		\put(20,-1.5){\small (a) overall network architecture}
		\put(69,-1.5){\small (b) operations between two layers of decoder}
	\end{overpic}
	\caption{\textbf{\netname structure}. \netname contains a sparse CNN encoder that extracts features from the input 3D points discretized into sparse voxels and three transformer decoders that generate vertex corners, edge curves and face patches respectively.
		The tri-path decoder interchanges data across different element groups throughout the layers to enable the emergence of a consistent B-Rep chain complex structure. \netname is trained with unary loss terms applied on $\widetilde{\mathbf{V}}, \widetilde{\mathbf{E}}, \widetilde{\mathbf{F}}$, \etc and binary loss terms applied on $\widetilde{\mathbf{FV}}, \widetilde{\mathbf{EV}}, \widetilde{\mathbf{FE}}$, which measure the differences from GT primitives and their mutual topology, respectively. 
		On the right, the specific circuit containing attention modules between two layers of the tri-path decoder is shown (\cf  (\ref{eq:decode_step1})-(\ref{eq:decode_step4})).}
	\label{fig:network}
\end{figure*}
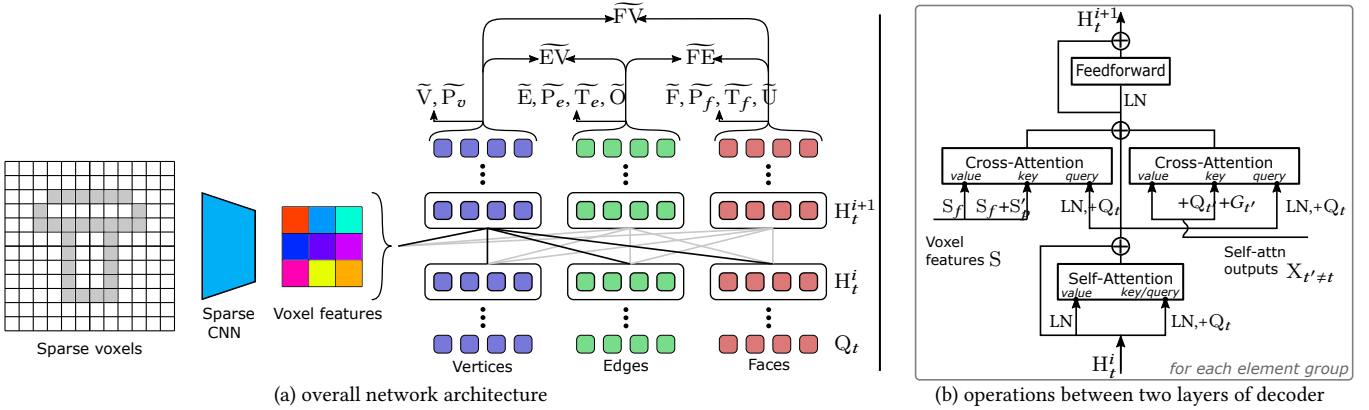

\subsection{Method Overview}

We formulate the problem of CAD reconstruction as B-Rep chain complex generation: given an unstructured point cloud $P=\{\mathbf{p}_i\in \mathbb{R}^3\}$ that captures the shape of the CAD model, we generate the chain complex $\mathcal{C}$ by detecting the primitive vertices $\mathbf{V}$, edges $\mathbf{E}$ and faces $\mathbf{F}$, their mutual relationships $\mathbf{FE}, \mathbf{EV}, \mathbf{FV}$, and their spatial embeddings through geometric primitives $\mathcal{P}$.

To obtain even a rough collection of geometric primitives and their relationships from an unstructured point cloud is a challenging task.
We design the \netname neural network to parse the point cloud with sparse CNN and reconstruct a whole complex through three paths of transformer decoders generating the elements and their mutual topology relationships (Fig.~\ref{fig:pipeline}(a)-(b) and Fig.~\ref{fig:network}).
In Sec.~\ref{sec:detection} we present the learning based framework.

While robust to different shape variations, the output of \netname is generally not a precise or valid B-Rep chain complex.
We treat the network predictions as a posterior probability on the occurrence of primitive elements and their correspondences, and finalize the extraction of a valid chain complex by global optimization that maximizes the likelihood while always satisfying topological constraints and fits the primitives to both input points and each other as prescribed by the explicit topology (Fig.~\ref{fig:pipeline}(b)-(d)).
In Sec.~\ref{sec:postproc} we present the global optimization procedure that enforces the above structural constraints and extracts valid B-Rep CAD models.

In Sec.~\ref{sec:results} we validate the design choices and compare with previous works through extensive tests on large scale datasets, and evaluate the robustness of our \name framework on stress tests including noisy and partial point clouds.

\section{Learning to generate B-Rep chain complexes}
\label{sec:detection}

We design a deep neural network to analyze the input point cloud and output a B-Rep chain complex structure.
The B-Rep structure is encoded by vertices $\widetilde{\mathbf{V}}$, edges $\widetilde{\mathbf{E}}$ and faces $\widetilde{\mathbf{F}}$, their mutual relationships $\widetilde{\mathbf{FE}}, \widetilde{\mathbf{EV}}, \widetilde{\mathbf{FV}}$, and their spatial embeddings through geometric primitives $\mathcal{P}$.
We use the tilde notation to denote the relaxation of binary variables to real numbers within the unit interval, for a probabilistic interpretation that guides the global optimization finalizing valid B-Rep structures (Sec.~\ref{sec:postproc}). 

\subsection{\netname Design}
\label{subsec:network}

The main challenge with network design is to enable the generation of a collection of elements with categorical topological relationships.
We use a SparseCNN as the encoder that extracts spatial features from the point cloud and transformer decoders to output the sets of corners, curves and patches as well as their mutual topology (Fig.~\ref{fig:network}).
To facilitate the emergence of consistent structures, we further introduce mutual communication between element groups in the transformer decoders, which effectively turns other groups of elements into a context for generating the current group.

\paragraph{SparseCNN encoder}
The encoder is a sparse convolutional neural network implemented with \cite{choy20194d}.
It takes as input the point set $P=\{\mathbf{p}_i\in \mathbb{R}^3\}$ with the spatial coordinates as 3-channel features for each point; point normal vectors are optionally provided.
The points are first discretized by a sparse grid of $128^3$ resolution and then convoluted and down-sampled by pooling gradually into a sparse grid of resolution $16^3$, where each non-empty voxel has a feature vector of size $d=384$.
We denote the output non-empty voxels as a pair of grid indices and feature vectors, i.e., $\mathbf{S} = [\mathbf{S}_p \in \mathbb{R}^{S\times 3}, \mathbf{S}_f \in \mathbb{R}^{S\times d}]$, with $S$ the number of non-empty voxels. 
The encoder output will be used by the decoder for element generation.
The complete network structure is provided in Appendix~\ref{appn:network_details}.

\paragraph{Tri-path transformer decoder}

As shown in Fig.~\ref{fig:network}, the decoder is a transformer-based architecture that decides for each of three element groups whether an element is valid (or present) in the input point cloud, and if valid, how it is connected by topology to other elements and how it embeds geometrically as a primitive shape.

The decoder takes as context $\mathbf{S}_p' = [\textrm{PE}(\mathbf{S}_p[i])]_{i\in[S]} \in \mathbb{R}^{S\times d}$ and $\mathbf{S}_f$ transformed from the voxels of encoder output, where $\textrm{PE}(\cdot)$ is the sinusoidal positional encoding \cite{transformerNIPS2017} that maps each of the three grid coordinates into a 128-dim vector and concatenates them into a 384-dim positional embedding.
Following \cite{DETR2020}, the sets of elements are initialized as learned embeddings, which we denote as $\mathbf{Q}_v{\in}\mathbb{R}^{N_v\times d}, \mathbf{Q}_e{\in}\mathbb{R}^{N_e\times d}, \mathbf{Q}_f{\in}\mathbb{R}^{N_f\times d}$ for vertices, edges and faces, respectively.

Given the context and element embeddings, the transformer decoder goes through three paths simultaneously that generate the three groups of elements representing a consistent B-Rep chain complex.
At each layer of the decoder, an element group first uses self-attention to exchange data inside the group, and then uses cross-attention to fetch contextual data from both the encoder output and the other groups of elements, which facilitates the learning of coherent B-Rep structures (\cf Sec.~\ref{sec:ablation}).
To distinguish the different element groups during cross-attention, we use the learned group embeddings $[\mathbf{G}_v,\mathbf{G}_e,\mathbf{G}_f]\in \mathbb{R}^{3\times d}$ corresponding to the vertex, edge and face types, respectively.

The specific update procedure between two decoder layers is given by (also illustrated in Fig.~\ref{fig:network}, right):
\begin{align}
	\mathbf{H}^{i}_t \pluseq\ &\textrm{SA}\!\left(\textrm{LN}(\mathbf{H}^{i}_t) {+} \mathbf{Q}_t, \textrm{LN}(\mathbf{H}^{i}_t)\right),\  \mathbf{X}_t = \textrm{LN}(\mathbf{H}^{i}_t), \label{eq:decode_step1}\\
	\mathbf{H}^{i}_t \pluseq\ &\textrm{CA}\!\left(\mathbf{X}_t {+} \mathbf{Q}_t, [\mathbf{X}_{t'\neq t} {+} \mathbf{Q}_{t'} {+} \mathbf{G}_{t'}], [\mathbf{X}_{t'\neq t}] \right), \label{eq:decode_step2}\\
	\mathbf{H}^{i}_t \pluseq\ &\textrm{CA}\!\left(\mathbf{X}_t {+} \mathbf{Q}_t, \mathbf{S}_f {+} \mathbf{S}_p', \mathbf{S}_f \right), \label{eq:decode_step3}\\
	\mathbf{H}^{i+1}_t =\ &\mathbf{H}^{i}_t + \textrm{FFN}\!\left(\textrm{LN}(\mathbf{H}^{i}_t)\right), \label{eq:decode_step4}
\end{align}
where $t,t'\in \{f,e,v\}$ iterate over the three element types, i.e., vertices, edges and faces, and $0\leq i \leq l-1$ iterates over the $l$ layers.
The network variables and modules are defined as follows: $\mathbf{H}^{i}_t\in \mathbb{R}^{N_t\times d}$ is the latent code at the $i$-th layer, initialized as $\mathbf{H}^{0}_t = \mathbf{Q}_t$,  $\mathbf{X}_t$ are intermediate tensors, $\textrm{LN}(\cdot)$ the layer normalization operation, $\textrm{SA}(\cdot,\cdot)$ the self-attention module whose first argument is the query/key tensor and second argument the value tensor, $\textrm{CA}(\cdot,\cdot,\cdot)$ the cross attention module with three arguments corresponding to query, key and value respectively, and finally $\textrm{FFN}(\cdot)$ the feed-forward module.
The addition between $\mathbf{Q}_t$ and $\mathbf{G}_t$ is done by broadcasting $\mathbf{G}_t$ to the first $N_t$ dimensions of $\mathbf{Q}_t$.

We have used a pre-normalization scheme where the latent codes are first normalized before being processing by attention modules and FFN ((\ref{eq:decode_step1}),(\ref{eq:decode_step4})), which has been shown to improve transformer training stability \cite{Xiong20}.
We also introduce the positional encodings and type embeddings for queries and voxels at each intermediate attention operation ((\ref{eq:decode_step1}),(\ref{eq:decode_step2}),(\ref{eq:decode_step3})), following and generalizing the detection framework of \cite{DETR2020}.

Given the final latent codes $\mathbf{H}_t^{l}$ for all element groups, we recover the attributes, including validness, geometric types and embeddings and topology, with specialized network modules, as presented next.

\paragraph{Validness prediction}
The validness module outputs the probability of an element being existent in the given input shape:
\begin{equation}
\textrm{F}^{val}_t: \mathbf{H}_t^{l}[i] \rightarrow \textrm{Prob}\left(\{\textit{valid}, \textit{invalid}\}\right),
\end{equation}
for each of the three element groups.
The module is implemented as three-layer MLPs, with details given in Appendix~\ref{appn:network_details}.
We denote the resulting validness probability as $\widetilde{\mathbf{V}}, \widetilde{\mathbf{E}}, \widetilde{\mathbf{F}}$ for vertices, edges and faces, respectively.
The element validness provides the base probability for all the rest classification tasks which further predict conditional probabilities over corresponding categorical domains.

\paragraph{Primitive type prediction}
The type modules, also implemented as three-layer MLPs, classify the geometric primitive types and other attributes of the elements. 

For edges we have
\begin{align}
&\textrm{F}^{type}_{e}: \mathbf{H}_e^l[i] \rightarrow \textrm{Prob}\left(\{\textit{line, circle, B-spline, ellipse}\}\right)\\
&\textrm{F}^{open}_{e}: \mathbf{H}_e^l[i] \rightarrow \textrm{Prob}\left(\{\textit{open, closed}\}\right),
\end{align}
which give the probability distributions for curve primitive types and open/closeness of each edge, respectively.
We denote the predicted edge types as $\widetilde{\mathbf{T}_e}$, and the openness probability as $\widetilde{\mathbf{O}}$.

For faces we have
\begin{align}
    \textrm{F}^{type}_{f}: \mathbf{H}_f^l[i] &\rightarrow  \\ \textrm{Prob}(&\{\textit{plane}, \textit{cylinder}, \textit{torus}, \textit{B-spline}, \textit{cone}, \textit{sphere}\})\\
    \textrm{F}^{open}_{f}: \mathbf{H}_f^l[i] &\rightarrow   \textrm{Prob}(\{\textit{u-open, u-closed}\}),
\end{align}
which give the probability distributions for patch primitive types and the open/closedness of a patch in one of its parametric dimensions.
We denote the predicted face types as $\widetilde{\mathbf{T}_f}$, and the u-openness probability as $\widetilde{\mathbf{U}}$.

Following the handling of parametric surfaces in \cite{ParseNet2020ECCV}, a patch instance is \textit{u-closed} if one of its parameter ranges forms a closed cyclic domain, examples including cylinder, cone, surfaces of revolution, torus, sphere, \etc that are closed in one dimension regarded as the u-dimension.
The \textit{u-closed} label of a groundtruth patch is used for enumerating the permutation patterns for geometry loss computation during training (Sec.~\ref{subsec:loss_functions}), and its prediction is used for closing up the resultant surface patch during the geometric refinement stage (Sec.~\ref{sec:geom_opt}).

\paragraph{Geometry prediction}

We generate the explicit geometry of elements by taking a unified representation for different types of edges and faces, in contrast to the online algebraic fitting to segmented points  \cite{ParseNet2020ECCV,LiPrimitive2019CVPR}.

We define the shape of an edge $e_i$ as a mapping from the canonical unit interval $[0,1]$ to a spatial curve, \ie $\mathcal{P}_e(\mathbf{H}_e^l[i]): [0,1] \rightarrow \mathbb{R}^3$, and the shape of a face $f_i$ as a mapping from the unit square $[0,1]^2$ to a spatial surface patch, \ie $\mathcal{P}_f(\mathbf{H}_f^l[i]): [0,1]^2 \rightarrow \mathbb{R}^3$.
The modules $\mathcal{P}_e, \mathcal{P}_f$ are implemented as hyper-networks \cite{MetaLearn2020,Hypernet2017}, where the latent codes of edges and faces are used to generate their specific three-layer MLPs, which then map the canonical parameter domains to spatial curves and surfaces. 
Fig.~\ref{fig:geom_embed} illustrates the mappings and network details are given in Appendix~\ref{appn:network_details}.
We discuss the choice of hypernet over an alternative conditioned MLP for implementing the mappings in the supplemental document.

Each curve is sampled uniformly from the unit interval with $K_e$ points, and each patch is sampled uniformly with $K_f\times K_f$ points.
We set $K_e = 30, K_f = 10$ in our implementation, and denote the sampled point arrays of all elements as $\widetilde{\mathbf{P}_e}, \widetilde{\mathbf{P}_f}$.

As the corner embedding generates a single point, we use a direct MLP instead, \ie $\mathcal{P}_v(\mathbf{H}_v^l[i]) \in \mathbb{R}^3$ for a vertex $v_i$.
The points of all corners are denoted as $\widetilde{\mathbf{P}_v}$.

We adopt the explicit geometry generation rather than online fitting, so that we can use the above unified module for geometry regression, which accommodates the different parametric forms (spanning both algebraic surfaces and freeform splines) and simplifies loss computation.
In comparison, ParseNet \cite{ParseNet2020ECCV} has to design a specific spline module pretrained to incorporate freeform surfaces, while \cite{LiPrimitive2019CVPR,ParseNet2020ECCV} solve nonlinear optimization problems during the network forward pass for primitive patch fitting excluding splines.

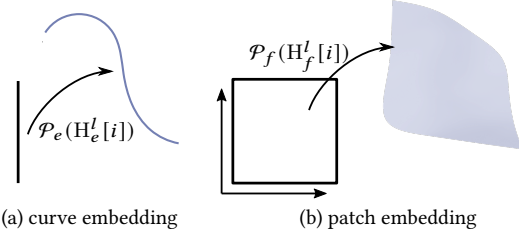
\begin{figure}
	\begin{overpic}[width=0.8\linewidth]{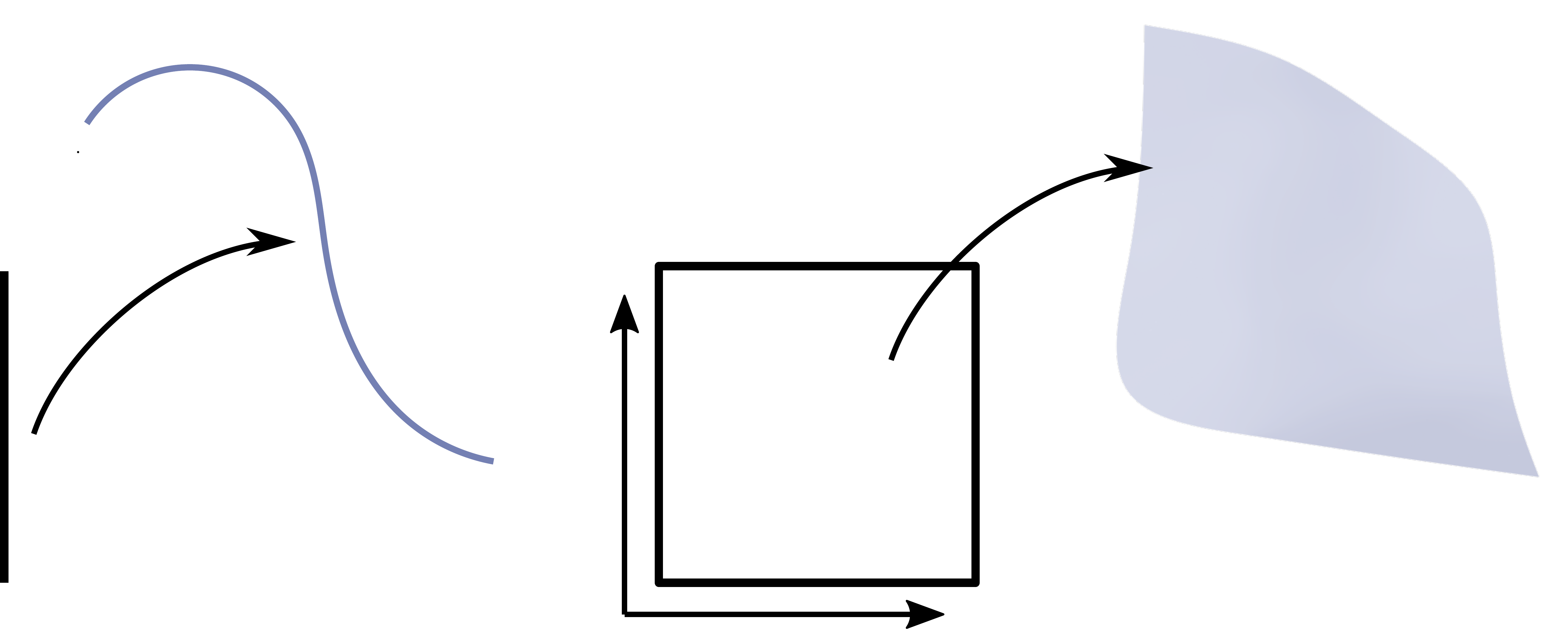}
		\put(-3,-5){\small (a) curve embedding}
		\put(4,12){\small $\mathcal{P}_e(\mathbf{H}_e^l[i])$}
		\put(55,-5){\small (b) patch embedding}
		\put(45,27){\small $\mathcal{P}_f(\mathbf{H}_f^l[i])$}
	\end{overpic}
	\vspace{0.5mm}
	\caption{\textbf{Geometry embedding of curves and patches}. Each embedding module is implemented as a hypernet that takes a latent code of an element and produces a mapping from the canonical domain (unit interval for curves, unit square for patches) to the spatial shapes. }
	\label{fig:geom_embed}
\end{figure}

\paragraph{Topology prediction}

For modeling topology, three modules $\textrm{F}^{topo}_{fe}$, $\textrm{F}^{topo}_{ev}$, $\textrm{F}^{topo}_{fv}$ output the respective probabilistic adjacency matrices.
For example, $\textrm{F}^{topo}_{fe}: \mathbf{H}_{f}^{l}, \mathbf{H}_{e}^{l} \rightarrow \widetilde{\mathbf{FE}} \in [0,1]^{N_f\times N_e}$, by first projecting each face(edge) element with a linear layer, and then computing the correlation of each pair of face/edge through their dot product followed by a sigmoid function $\sigma$, \ie
\begin{equation}
	\textrm{F}^{topo}_{fe}(\mathbf{H}_{f}^{l}, \mathbf{H}_{e}^{l}) = \left[ \sigma\left(\textrm{FC}_{fe}\left(\mathbf{H}_{f}^{l}[i]\right) \cdot  \textrm{FC}_{ef}\left(\mathbf{H}_{e}^{l}[j]\right)\right) \right]_{i,j},
\end{equation}
for $i\in [N_f], j\in [N_e]$.
The $\textrm{F}^{topo}_{ev}, \textrm{F}^{topo}_{fv}$ modules are defined similarly.
Such a design captures the pairwise correlation that the discrete topology models.
While the topology predictions are fitted to sparse adjacency matrices, the dot-product formulation provides no short-cut to introduce imbalance bias.

From ablation studies (Sec.~\ref{sec:ablation}), it is shown that compared with an implicit encoding of topology through geometric proximity, the explicit modeling of topology provides a different source of information that complements pure geometrical data to enhance structurally valid B-Rep reconstruction.
For example, topology of cluttered elements is ambiguous through their geometric proximity only, but can be sharply encoded by the predicted adjacency matrices. 

\subsection{Loss Functions}
\label{subsec:loss_functions}

The ground truth B-Rep structure contains element sets of diverse sizes without canonical ordering. 
Given the predicted element sets, we compare them with the ground truth data by first building a matching between corresponding sets and then defining loss functions based on the matched pairs.

\paragraph{Matching}

We build a correspondence between the GT elements and the generated elements by bipartite matching \cite{Kuhn1955,Munkres1957}.
The cost matrix for matching is roughly the same as loss computation, so that the correspondences among consecutive training iterations are changed gradually to stabilize training.

In particular, for each pair of prediction element $p$ and ground truth element $q$, we compute the matching cost as:
\begin{equation}
	C(p,q) = \sum_{c}{D_{KL}\left(c(q)||c(p)\right)} + w_{geo} D_{geo}(p, q),
\end{equation}
where $c$ iterates over all classification tasks applicable to the element (\ie validness, type, curve/patch openness) and computes the probability distributions,
$D_{KL}$ the KL divergence measures the negative log-likelihood of $p$ equal to $q$ on each task,
$w_{geo}=300$ gives a larger weight to geometric error as the shapes are normalized into a unit cube, 
and $D_{geo}\in \{ D_v, D_e, D_f \}$ computes the geometric distance between two elements, as defined next.

For each element group, the matching process divides the generated elements into two sets, the matched ones $M_t$ and the unmatched;
we denote the correspondence of matched elements to GT elements as $m(\cdot)$ and the inverse correspondence to prediction as $m'(\cdot)$.
We have chosen the element group sizes $N_{v/e/f}$ to cover GT elements for all samples in the dataset (Sec.~\ref{sec:results}).

\begin{remark}
The matching process considers per-element attributes only for cost computation, while ideally it should also include the binary topology prediction.
However, we note that by taking binary terms into consideration, the matching task becomes a quadratic assignment problem that has no efficient solution \cite{Anstreicher2003QAP}.
Instead of solving an intractable quadratic matching problem during each training iteration, the topology loss and tri-path network structure enable the co-occurrence of consistent elements, as shown empirically in Sec.~\ref{sec:ablation}.
\end{remark}

\paragraph{Loss terms}
The loss functions measure differences from ground-truth labels in the following aspects: detection validness, primitive types, geometric regression and topology between every pair of element groups. 
Among them, the geometric regression loss measures distance from ground-truth shapes, and the rest computes cross entropy from target categorical distributions.
We denote the predicted variables as $\widetilde{\mathcal{C}}$ with tilde, and the GT complex as $\overline{\mathcal{C}}$.

\paragraph{Validness}
The validness loss encourages the matched elements of prediction be valid and the rest be invalid.
Thus we have the following binary classification problem for all generated elements:
\begin{align}
    L_{val} &= \frac{1}{N_v}\sum_{i=1}^{N_v}{\textrm{BCE}(\widetilde{\mathbf{V}}[i], \mathbf{1}_{i\in M_v})} + \frac{1}{N_e}\sum_{i=1}^{N_e}{\textrm{BCE}(\widetilde{\mathbf{E}}[i], \mathbf{1}_{i\in M_e})} \nonumber\\ 
    &+ \frac{1}{N_f}\sum_{i=1}^{N_f}{\textrm{BCE}(\widetilde{\mathbf{F}}[i], \mathbf{1}_{i\in M_f})},
\end{align}
where $\textrm{BCE}(\cdot,\cdot)$ computes the binary cross entropy. $\mathbf{1}_{x}$ is the indicator function that is 1 when $x$ is true and 0 otherwise.

\paragraph{Type classification}
The primitive type classification loss compares the types of generated elements with those of the ground truth elements, for edges and faces. 
The edges also have the type of being open/closed, while faces have the type of being u-open/u-closed.
For type classification, we only supervise the matched elements:
\begin{align}
	L_{cls} =& \frac{1}{|M_e|}\sum_{i\in M_e}{\textrm{CE}\left(\widetilde{\mathbf{T}_e}[i], \overline{\mathbf{T}_e}[m(i)]\right) + \textrm{BCE}\left(\widetilde{\mathbf{O}}[i], \overline{\mathbf{O}}[m(i)]\right)} \nonumber \\
	 + \frac{1}{|M_f|}&\sum_{i\in M_f}{\textrm{CE}\left(\widetilde{\mathbf{T}_f}[i], \overline{\mathbf{T}_f}[m(i)]\right) + \textrm{BCE}\left( \widetilde{\mathbf{U}}[i], \overline{\mathbf{U}}[m(i)] \right)},
\end{align}
where $\textrm{CE}(\cdot,\cdot)$ computes the cross entropy between two probability distributions, $\mathbf{T}$ denotes the primitive type probabilities, and $\mathbf{U}$ denotes patch u-open/u-closed probabilities.

\paragraph{Geometry regression}
Geometry regression of the generated elements toward matched GT elements is supervised by:
\begin{align}
	L_{geo} = &\frac{1}{|M_v|}\sum_{i\in M_v}{D_v\left(\widetilde{\mathbf{P}_v}[i], \overline{\mathbf{P}_v}[m(i)] \right)} \nonumber\\
	&+ \frac{1}{|M_e|}\sum_{i\in M_e}{D_e\left(\widetilde{\mathbf{P}_e}[i], \overline{\mathbf{P}_e}[m(i)]\right)} \nonumber\\
	&+ \frac{1}{|M_f|}\sum_{i\in M_f}{D_f\left(\widetilde{\mathbf{P}_f}[i], \overline{\mathbf{P}_f}[m(i)]\right)},
\end{align}
where $D_v(\mathbf{p}, \mathbf{q}) = \left\| \mathbf{p} - \mathbf{q} \right\|^2$ computes the squared distance between two vertices. 
$D_e(\cdot,\cdot)$ is the curve distance that chooses the minimum distance among all sequential 1-1 matches of two linear arrays of curve points. 
$D_f(\cdot,\cdot)$ is the patch distance that compares a pair of two-dimensional arrays of patch points with sequential 1-1 matches preserving array ordering.

Specifically, for two curves $\mathbf{P}_0, \mathbf{P}_1 \in \mathbb{R}^{K_e\times 3}$ of length $K_e$, we define the edge distance as 
\begin{equation}
	D_e(\mathbf{P}_0, \mathbf{P}_1) = 
	\frac{1}{K_e}	\min_{\Pi} \|\mathbf{P}_0 - \Pi(\mathbf{P}_1)\|^2.
\end{equation}
If $\mathbf{P}_1$ is an open curve, $\Pi\in \{\textrm{rev}|\cdot\}$ enumerates the 2 possible ways of flipping the curve, with reversion by $\textrm{rev}$ or not.
If $\mathbf{P}_1$ is a closed curve, $\Pi\in \{\textrm{roll}_i\circ(\textrm{rev}|\cdot)\}$ additionally enumerates rolling the curve point array cyclically by $i$ entries, $0\leq i < K_e$.

For two patches $\mathbf{P}_0, \mathbf{P}_1 \in \mathbb{R}^{K_f\times K_f\times 3}$ each with $K_f\times K_f$ points, we define the face distance as
\begin{equation}
	D_f(\mathbf{P}_0, \mathbf{P}_1) = \frac{1}{K_f^2} \min_{\Pi}\|\mathbf{P}_0 - \Pi(\mathbf{P}_1)\|^2.
\end{equation}
When $\mathbf{P}_1$ is an open patch, $\Pi \in \{(\textrm{rev-x}|\cdot)\circ(\textrm{rev-y}|\cdot)\}$ enumerates all possible flipping of a 2D grid, i.e. reversing (or not) along the x/y axis in 4 possible ways.
On the other hand, when $\mathbf{P}_1$ is a closed patch, $\Pi \in \{\textrm{roll-y}_j\circ\textrm{roll-x}_i\circ(\textrm{rev-x}|\cdot)\circ(\textrm{rev-y}|\cdot)\}$ additionally enumerates all possible shifting of a 2D grid, i.e. rolling along the x/y axis for steps within $K_f$.

The reason for using order preserving 1-1 matching distances for curves and patches is to learn more regular shapes; in comparison, aggregated metrics like chamfer distance do not penalize shape irregularity and lead to poor generation quality.
A similar distance is used for regular spline patch regression in \cite{ParseNet2020ECCV}.
In addition, we prepare the ground-truth patches by regular grid sampling of their parametric representations, as is done in \cite{ParseNet2020ECCV} for spline regression training.

\paragraph{Topology}

Topology prediction is compared with the ground-truth adjacency topology for the matched elements:
\begin{align}
	L_{topo} = &\frac{1}{|M_e||M_v|}\sum_{i\in M_e, j\in M_v}\textrm{BCE}\left(\widetilde{\mathbf{EV}}[i,j], \overline{\mathbf{EV}}[m(i), m(j)] \right) \nonumber\\
	& + \frac{1}{|M_f||M_e|}\sum_{i\in M_f, j\in M_e}\textrm{BCE}\left(\widetilde{\mathbf{FE}}[i,j], \overline{\mathbf{FE}}[m(i), m(j)] \right) \nonumber \\
	& + \frac{1}{|M_f||M_v|}\sum_{i\in M_f, j\in M_v}\textrm{BCE}\left(\widetilde{\mathbf{FV}}[i,j], \overline{\mathbf{FV}}[m(i), m(j)] \right).
\end{align}

\paragraph{Total loss function}
The final loss function used for training the whole network is a weighted summation of the above terms:
\begin{equation}
	L = L_{val} + L_{cls} + w_{geo} L_{geo} + w_{topo} L_{topo}. %+ L_{cons}
\end{equation}
Since the models are normalized into unit side-length bounding boxes, we have used $w_{geo}=300$ and $w_{topo}=10$ for all experiments.

\begin{remark}
Due to the full supervision of both geometry and topology predictions in this task, we do not include the consistency of geometry and topology nor the topology constraints ((\ref{eq:opt:cons:manifold})-(\ref{eq:opt:cons:closed})) into the objective function.
However, we note that these additional targets can be used for self-supervision when ground-truth labels are missing on other datasets.
\end{remark}

\section{Neurally guided B-Rep reconstruction}
\label{sec:postproc}

Given the fuzzy predictions of the geometric elements and their mutual topology, we apply global optimization to recover both the primitive geometries and their topology that constitute a valid B-Rep chain complex.
The process is done in two steps: the chain complex extraction that recovers valid topological structures (Sec.~\ref{sec:topo_opt}) and the geometry refinement that realizes the topological structures and fitness to input points (Sec.~\ref{sec:geom_opt}).

\subsection{Chain Complex Extraction}
\label{sec:topo_opt}

The aim of complex extraction is to recover a structurally valid manifold B-Rep model as dictated by ((\ref{eq:opt:cons:manifold})-(\ref{eq:opt:cons:closed})), while obeying the predictions given by the network on the posterior topology and geometry as much as possible.

To be precise, we search for the corresponding binary variables $\mathbf{V}, \mathbf{E}, \mathbf{O}, \mathbf{F}$ and $\mathbf{FE}, \mathbf{EV}, \mathbf{FV}$ that represent a valid topological structure and agree with the predicted likelihoods and geometric proximity as much as possible.

Denote the predictions in element validness as $\widetilde{\mathbf{V}}, \widetilde{\mathbf{E}}, \widetilde{\mathbf{O}}, \widetilde{\mathbf{F}}$, and their mutual relationships as $\widetilde{\mathbf{FE}}, \widetilde{\mathbf{EV}}, \widetilde{\mathbf{FV}}$.
As the predicted topology models conditional probability (Sec.~\ref{subsec:network}), we multiply the validness probability to the predicted topology matrices to obtain the true probabilities of topology:
\begin{equation}
     \widetilde{\mathbf{FE}}[i,j] := \widetilde{\mathbf{FE}}[i,j]\times \widetilde{\mathbf{F}}[i]\times \widetilde{\mathbf{E}}[j], 
\end{equation}
and similarly for $\widetilde{\mathbf{EV}}, \widetilde{\mathbf{FV}}$.

We further compute the geometric proximity of elements that provides a fitness measure of possible topology connections.
Let
\begin{equation}
d_{a,b} = \frac{1}{|P_a|}\sum_{\mathbf{p}_a\in P_a}{\min_{\mathbf{p}_b\in P_b} \|\mathbf{p}_a - \mathbf{p}_b\|}
\label{eq:adj_dist}
\end{equation}
be the average distance between the predicted point sets of two elements $a,b$, with the order of $a$ lower than $b$ (\eg corner to curve).
We define the fitness score of two elements being connected as $S(a,b) = \exp(-\frac{d_{a,b}^2}{\epsilon^2}) \in (0,1]$, where $\epsilon=0.1$ controls the fall-off rate of the fitness score.
The adjacency probability matrices computed purely from geometry proximity are then given by 
\begin{equation}
 \mathbf{S}_{FE} = \left[S(e_j,f_i)\right]_{i,j}, \mathbf{S}_{EV} = \left[S(v_j,e_i)\right]_{i,j}, \mathbf{S}_{FV} = \left[S(v_j,f_i)\right]_{i,j}.
 \label{eq:geom_prox_topo}
\end{equation}

\paragraph{Non-maximum suppression}
Before going through a combinatorial optimization that searches for the valid structure based on network predictions, we remove some apparently redundant elements by a simple nonmaximum suppression step that removes duplicate elements.
We define a duplicate element $q$ as: 1) it is predicted to be valid (i.e. its validness $\geq 0.5$), and 2) there exists $q'$ such that $q'$ has validness $\geq 0.5$, the same type as $q$, the same topology w.r.t other element groups as $q$, and nearly the same geometry, computed by their chamfer distance within a given threshold (we use $0.05$ for all experiments).
For a duplicate element, we simply set its validity and corresponding rows and columns in the topology matrices to zero.

\paragraph{Combinatorial optimization}
We formulate a binary optimization problem with linear objectives to extract the complex structure:
\begin{align}
	\textbf{max} &\quad w F_{topo} + (1-w) F_{geom} \label{eq:chain_extraction_problem} \\
	\textbf{s.t.} &\quad (\ref{eq:opt:cons:manifold}), (\ref{eq:opt:cons:curveend}), (\ref{eq:opt:cons:closed}) \nonumber \\
	 &\quad \begin{cases} 
		        \mathbf{FE}[i,j] \leq \mathbf{F}[i] \left(\leq \sum_{j}{\mathbf{FE}[i,j]}\right), &\forall i,j\\
		        \mathbf{EV}[i,j] \leq \mathbf{V}[j] \leq \sum_{k}{\mathbf{EV}[k,j]}, &\forall i,j
	        \end{cases} \nonumber 
\end{align}
where $w=0.5$ is the weight given to the topology fitness objective.

The constraints come from B-Rep manifoldness (Sec.~\ref{sec:brep_formulation}) and the dependency between binary topology terms and unary terms denoting the existence of elements.
The dependency between binary variables and unary variables is necessary.
Intuitively, for example, the absence of a face dictates the absence of its adjacency with edges, and due to Eq.~(\ref{eq:opt:cons:closed}) the absence of adjacency with vertices;
on the other hand, when a face exists, it generally requires adjacent edges and vertices as boundaries, except the singular case of all surface patches are predicted closed and no edges are predicted existent, hence the parenthesis in the first inequality constraint above.
We find that all such dependencies can be derived from the constraints formulated in the above problem, a proof of which is provided in the supplemental material.

We define the topology objective as 
\begin{equation}
	F_{topo} = \sum_{\mathbf{X}}{w_{X}\left(2\widetilde{\mathbf{X}}-1\right) \cdot \mathbf{X}},
\end{equation}
where $\mathbf{X}$ ranges over the variables $\{ \mathbf{F}, \mathbf{E}, \mathbf{V}, \mathbf{O}, \mathbf{FE}, \mathbf{EV}, \mathbf{FV} \}$, $\widetilde{\mathbf{X}}$ over the corresponding predictions, and $\cdot$ denotes the dot product between two vectors or matrices.
We assign a larger weight $w_X=10$ for the unary variables $\{ \mathbf{F}, \mathbf{E}, \mathbf{V}, \mathbf{O}\}$, and a unit weight $w_X=1$ for the binary variables $\{\mathbf{FE}, \mathbf{EV}, \mathbf{FV} \}$, to balance their significantly different numbers of variables.
The $2x - 1$ coefficient is to encourage those variables with likelihood larger than 0.5 to emerge and those less than 0.5 to turn off.

Similarly we define the geometry fitness objective with proximity induced adjacency likelihoods as
\begin{equation}
	F_{geom} = \sum_{\mathbf{X}}{w_{X}\left(2\mathbf{S}_{X}-1\right) \cdot \mathbf{X}}
\end{equation}
where $\mathbf{X}$ ranges over $\{\mathbf{FE}, \mathbf{EV}, \mathbf{FV}\}$, and $\mathbf{S}_{X}$ defined in (\ref{eq:geom_prox_topo}).

The above optimization problem is hard to solve, because of the large number of binary variables as well as the quadratic constraints (\ref{eq:opt:cons:curveend}), (\ref{eq:opt:cons:closed}).
To make the problem more approachable, we reduce the binary variables significantly by simply removing the candidate vertices/edges/faces with predicted validness less than $0.3$, and convert the quadratic constraints to linear ones by a trick for binary variables.
The resultant integer linear program (ILP) is much more tractable and can be solved by state-of-the-art optimizers (we use \cite{gurobi} in experiments).
More details about the conversion are given in Appendix~\ref{appn:struct_extraction_detail}.
In practice, we set a time limit of 20 minutes for each ILP problem, and find that all 3k samples except one have been successfully solved.
The only failure case has low curve validness values truncated to zero by the 0.3 threshold and when we relax the threshold to 0.1, it can be successfully solved.

\subsection{Geometric Refinement}
\label{sec:geom_opt}

\begin{algorithm}[tb]
	\SetAlgoLined 
	\KwIn{complex with solved topology and predicted geometric primitives, input points, $K_1=3, K_2=5$} 
	\KwOut{geometric primitives conforming to prescribed topology and input points}
	
	\tcp{First stage: fitting with spline patches}
	\For{$i\in [K_1]$}{
		fit each patch as spline (except plane, sphere) to input points, curves, and corners\;
		fit each curve to patches and corners\;
		fit each corner to curves and patches\;
	}
	
	\tcp{Second stage: fitting with typed primitive patches}
	convert spline patches to typed primitives\;
	\For{$i\in [K_2]$}{
		fit each typed patch to input points, curves, and corners\;
		fit each curve to patches and corners\;
		fit each corner to curves and patches\;
	}
	\caption{Procedure of geometric refinement } 
	\label{alg:geometric_realization}
\end{algorithm}

Given the solved structure, we refine the geometry of the elements by fitting their corresponding typed geometric representations to input points, as well as constraining their shapes so that the topological structures are tightly satisfied.
Curves (patches) which are solved (predicted) to be closed are constructed accordingly.

The geometric refinement process consists of iterative applications of standard surface, curve and point fitting procedures.
The input points and adjacent elements prescribed by topology are the fitting targets.
We give different weights to the targets: $1$ for input points, $5$ for adjacent elements and $0.1$ for a stabilization term that prevents the element from drifting too far.
Note that only patches are fitted to the input points, as it is highly ambiguous to find the proper corresponding points for curves and corners.
To obtain the target points for a patch, we project the input points to their closest surface patches (initialized by predicted geometry), and filter out those with distances above a given threshold $0.02$ to make the fitting more robust to inaccurate initialization.

As shown in Alg.~\ref{alg:geometric_realization}, we separate the geometric refinement process into two stages.
In the first stage we use spline patches for fitting all types of primitive patches except plane and sphere, since we find that the fitting of cone and cylinder is highly sensitive to the normal vectors of assigned points, while spline fitting does not rely on the normal vectors.
In the second stage, we turn each primitive patch to its predicted type and apply the iterative fitting process again with type constraints; the fitted spline patches from the first stage provide high quality initial estimates of shapes and normal vectors for challenging primitives like cone and cylinder.

Each fitting problem is solved with standard techniques.
For types like lines, circles, planes, spheres, torus and splines, we use (nonlinear) least square fitting procedures \cite{EberlyGeometricTools}.
For types with quadratic algebraic equations, we adopt the methods of \cite{LiPrimitive2019CVPR,Andrews:2013:AQD} but apply additional constraints, \eg on cone and cylinder axes, as discussed next.

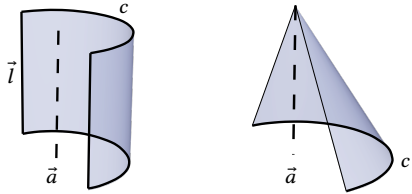
\begin{figure}
    \centering
    \begin{overpic}[width=0.6\linewidth]{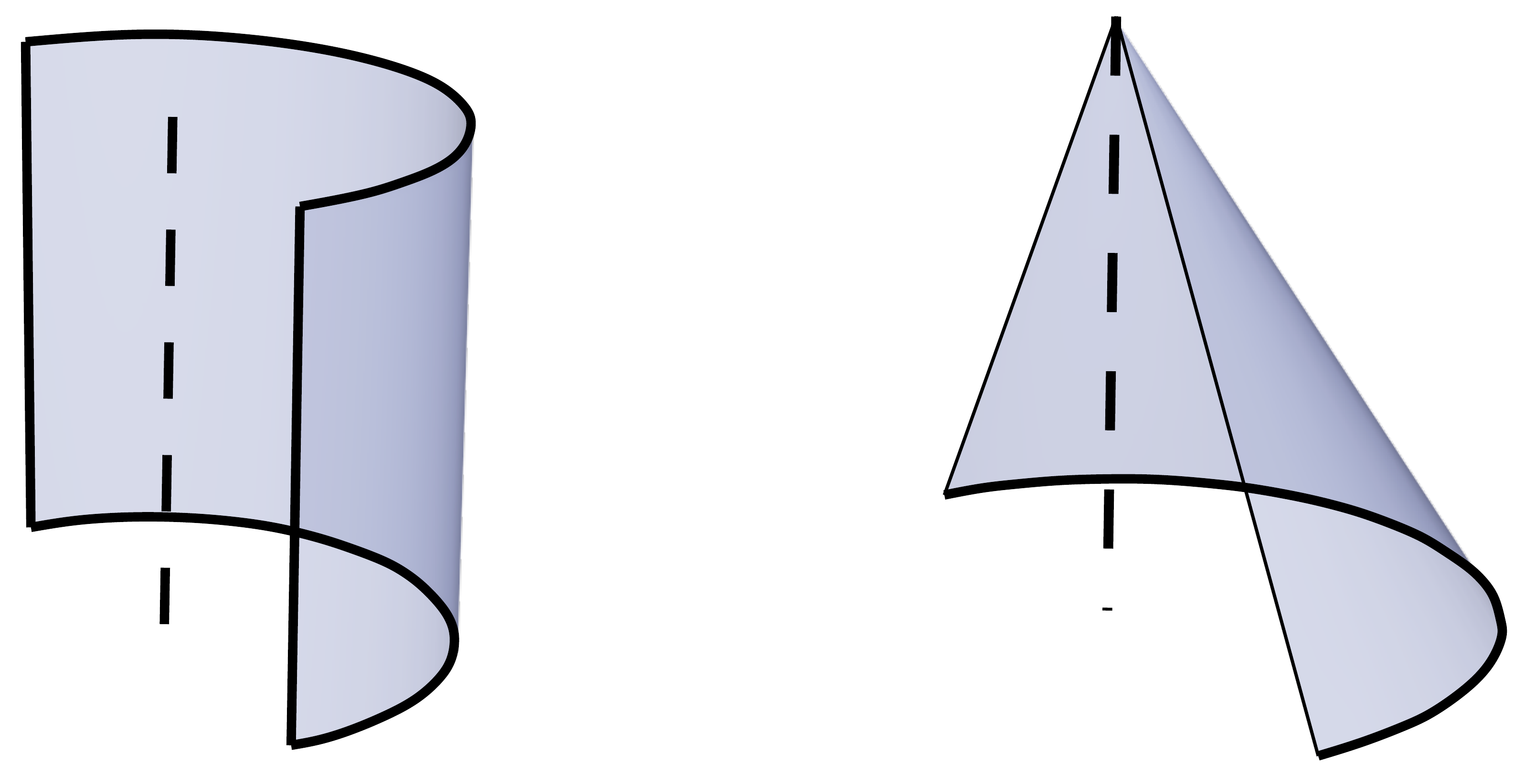}
        \put(8,4){\small $\vec{a}$}
        \put(-2,30){\small $\vec{l}$}
        \put(27,48){\small $c$}
        \put(70,4){\small $\vec{a}$}
        \put(100,8){\small $c$}
    \end{overpic}
    \caption{\textbf{Topology provides constraints for geometric fitting.} Left: the cylinder axis $\vec{a}$ is constrained to be parallel to linear boundary $\vec{l}$ and perpendicular to the circular boundary $c$. Right: the cone axis $\vec{a}$ is constrained to be perpendicular to the circular boundary $c$. }
    \label{fig:topo_cues}
\end{figure}

\paragraph{Topological cues for regularization.}
For challenging cases like narrow patches (see the pink narrow band in Fig.~\ref{fig:pipeline} for an example), traditional approaches that apply direct point cloud fitting become highly sensitive to outliers.
In contrast, our B-Rep representation provides additional cues for regularization.
For example, given a cylinder patch with four-sided boundary curves, if two curves are lines and the other two are circles in type, we instantly have that the axis of the cylinder is parallel to the lines and perpendicular to the circles. 
Similarly, for a cone patch, its axis can be determined from the circular boundary curves.
See Fig.~\ref{fig:topo_cues} for an illustration.
Since the estimation of axes of cones and cylinders is fundamental to their surface fitting and quite sensitive to estimated point normals \cite{Andrews:2013:AQD}, such regularization can be very helpful when the input points are noisy or highly incomplete.

We note that once the topological relationships are present, potentially more geometric constraints can be derived and applied for regularization. This inference and application of sophisticated constraints is generally known as \textit{auto-constraint}, an important feature in CAD modeling software \cite{autocad}.
Our reconstruction with explicit and definite topological relationships provides a foundation for such applications \cite{GlobFit2011,ZoneGraph2021,ConstraintSynthesis2021}, which we leave as future work to explore.

\paragraph{Model visualization}

\begin{figure*}[t]
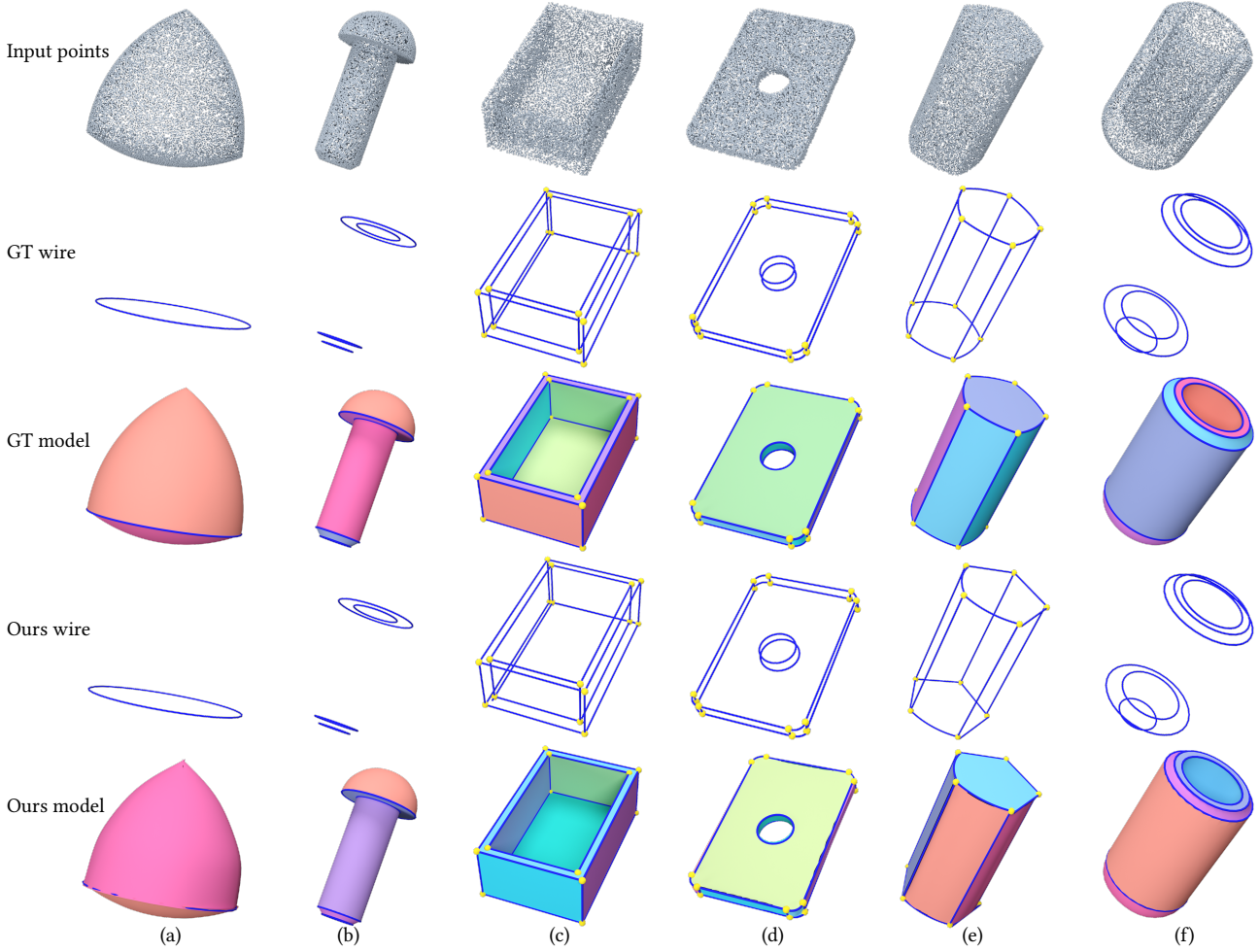

    \centering
    \begin{overpic}[width=0.9\linewidth]{image/gallery_revision_2}
        \put(-6,74){\small Input points}
        \put(-6,57){\small GT wire}
        \put(-6,41){\small GT model}
        \put(-6,25){\small Ours wire}
        \put(-6,10){\small Ours model}
        
        \put(7,-1){\small (a)}
        \put(22,-1){\small (b)}
        \put(40,-1){\small (c)}
        \put(58,-1){\small (d)}
        \put(75,-1){\small (e)}
        \put(93,-1){\small (f)}
    \end{overpic}
    \caption{\textbf{Result gallery}. Our framework can recover CAD models with complete B-Rep structures from the unstructured input points. Freeform surfaces (a)(e), smooth junction corners and curves (d) and narrow surface patches (b)(c)(d)(f) can be generated by our framework.}
    \label{fig:result_gallery}
\end{figure*}

The optimized B-Rep model can be converted to specific formats for CAD software consumption.
To visualize the whole models, we develop a simple procedure that extracts mesh models from the B-Rep chain complex: we use curve loops to cut their incident patches triangulated and obtain a collection of trimmed patches.
The final patches, curves and corners form a mesh model that follows the predicted topology and fits to the input geometry.
Examples of diverse complexities are shown in Figs.~\ref{fig:teaser}, \ref{fig:result_gallery}, \ref{fig:comparison} and \ref{fig:noise_partial_data} and the supplemental document.

\subsection{Structure Validity Assessment}
\label{sec:validity_checking}

Although results obtained from network predictions inevitably have degenerate cases, previous works rarely address the question of whether the result models are valid CAD models, as the results lack structures to validate on.
However, with the separate encoding of topology and geometry and the guaranteed topology validness by constraints (Sec.~\ref{sec:brep_formulation}), our B-Rep chain complex formulation provides an approach for assessing the validness of reconstructed CAD models by checking topology and geometry consistency.

We validate if the final geometric realization sufficiently complies with the solved topology, by checking if the algebraic adjacency between elements is met with sufficient geometric accuracy.
Precisely, we compute the corner-to-curve, corner-to-patch, and curve-to-patch distances through (\ref{eq:adj_dist}) for all adjacent pairs designated by topology, and compute the percentage of distances exceeding a given threshold.
Empirical results are given in Sec.~\ref{sec:validness_results}.

\section{Results and discussion}
\label{sec:results}

Through extensive ablation studies, comparisons and stress tests with noisy and missing data, we show that the proposed ComplexGen framework that models B-Rep structure holistically achieves more complete and structured CAD reconstruction.

\subsection{Setup}

\paragraph{Dataset}
Following previous work \cite{ParseNet2020ECCV}, we build a dataset using a portion of the ABC dataset \cite{Koch2019ABC} that contains 20k models for training and 3k models for testing, with the same split as \cite{ParseNet2020ECCV} for fair comparisons.
We apply data cleaning to remove errors and reduce label ambiguity; details of the procedure can be found in the supplemental document.
We make the fixed dataset publicly available to facilitate future research.

The B-Rep chain complex structure can be easily obtained for a sample in the ABC dataset.
To this end, different from \cite{ParseNet2020ECCV}, we also need to extract the corners and curves and the adjacency among elements, by cross-referencing the CAD feature file for element definition and the mesh representation that discretizes the CAD model for their adjacency.

We use $N_v = 100$, $N_e = 150$, $N_f = 100$ for the three element groups, which cover all GT elements in the dataset.
The input points have normal vectors for most experiments, except the stress tests with noisy data where we remove normal vectors to simulate a more practical and challenging setting.

\paragraph{Network training}
We use Adam solver \cite{AdamSolver} with default parameters and a fixed learning rate $lr=10^{-4}$ to train the \netname implemented in PyTorch with batchsize 48 for 500 epochs to convergence, which takes 3 days on 8 Nvidia V100 GPUs.

\paragraph{Runtime}
The network inference takes 89 ms. The most time consuming part is chain complex extraction that solves integer linear programming, taking 545 seconds on average on the test set.
The geometric refinement step takes 8 seconds on average.

\subsection{Evaluation Metrics}
\label{sec:eval_metric}

To evaluate the reconstruction accuracy of a complete B-Rep chain complex against ground truth, we come up with a comprehensive set of metrics measuring the detection accuracy of corners, curves, patches, the classification accuracy of primitive types, as well as the topological errors from GT and from validness constraints.

Given the ground-truth structure $\overline{C}$ and the reconstructed structure $C$, we compute the linear assignment matching for each pair of element groups by the same procedure as in Sec.~\ref{subsec:loss_functions} (albeit using distance cost only, as the categorical probability distributions are rounded to discrete values).
Denoting the matched prediction elements as $M_t$, the correspondence to GT elements as $m(\cdot)$ and the inverse correspondence to prediction as $m'(\cdot)$, we evaluate the following metrics.
\begin{itemize}[leftmargin=10pt,itemsep=1pt]
    \item \textbf{Detection accuracy by F-score}.  We apply a geometric distance threshold $\delta=0.1$  to determine if a pair of matched elements is a true positive. Then we compute the F-score of the detection as the harmonic mean of the precision and recall values. 
    \item \textbf{Type accuracy}. For each type classification task (including primitive types and openness of curves and patches), we measure the type accuracy for matched prediction elements, defined as $\frac{1}{|M_t|}\sum_{i\in M_t}{\mathbf{1}_{T_i=\overline{T}_{m(i)}}}$, where $T_i,\overline{T}_{m(i)}$ are the predicted type and corresponding GT type, respectively.
    \item \textbf{Topology error}. Given the GT adjacency matrix $\overline{\mathbf{FE}}$, we compute the error of the reconstructed topology as its difference from the GT matrix, \ie
    {
   	\abovedisplayskip=1pt
   	\abovedisplayshortskip=1pt
   	\belowdisplayskip=1pt
   	\belowdisplayshortskip=1pt
    \begin{equation}
     \frac{1}{\overline{N_f}\cdot\overline{N_e}}\sum_{i\in[\overline{N_f}], j\in [\overline{N_e}]}{ \left| \overline{\mathbf{FE}}[i,j] - \mathbf{FE}[m'(i), m'(j)] \right| }.
    \end{equation}
	}%
    The topology errors for $\mathbf{FV}$ and $\mathbf{EV}$ are computed similarly.
    Note that for a GT element pair $i,j$ with either one unmatched to prediction, since the prediction $\mathbf{FE}$ cannot be evaluated, the corresponding error is simply count as 1, which avoids the degeneracy that fewer predictions unfavorably lead to lower errors.
    \item \textbf{Topology inconsistency}. We evaluate the degree of topology inconsistency as the absolute residual of the three systems of equations (\ref{eq:opt:cons:manifold}),(\ref{eq:opt:cons:curveend}),(\ref{eq:opt:cons:closed}), each normalized by its number of equations.
\end{itemize}%
These metrics are primarily used for ablation study that compares alternative approaches for complex generation, as discussed in Sec.~\ref{sec:ablation}.

For comparison with previous works \cite{LiPrimitive2019CVPR,ParseNet2020ECCV,yan2021hpnet} that focus on patch segmentation, we evaluate our results against their metrics that measure the accuracy and coverage of reconstructed surface patches, and propose a patch-to-patch topology accuracy metric for measuring structural fidelity.
\begin{itemize}[leftmargin=10pt,itemsep=1pt]
    \item \textbf{Residual}. The residual error measures how well each generated patch fits to its target segment points.
    For matched patches, the residual is computed as 
    \begin{equation}
        \frac{1}{|M_f|}\sum_{i\in M_f}{\frac{1}{|S_{m(i)}|}\sum_{\mathbf{p}\in S_{m(i)}} \|\mathbf{p} - \text{proj}(\mathbf{p}, P_i)\|}, 
    \end{equation}
    where $P_i$ is the predicted patch, $S_{m(i)}$ the corresponding GT patch sample points, and $\text{proj}(\cdot,\cdot)$ projects a point onto a surface patch.
    \item \textbf{P-coverage}. The point coverage measures the percentage of input points covered by the predicted patches, computed by
    \begin{equation}
        \frac{1}{|P|}\sum_{\mathbf{p}\in P}{ \mathbf{1}_{\|\mathbf{p} - \text{proj}(\mathbf{p}, \{P_i\})\| < \epsilon_{cov}}},
    \end{equation}
    where $\mathbf{p}$ iterates over input points $P$ and $\epsilon_{cov} = 0.01$ following previous works.
    \item \textbf{Patch-to-patch topology error}. Since only patches are generated for segmentation based methods, we define the patch-to-patch topology error in the same manner as the other topology error metrics defined above, to measure the structural fidelity to GT patches.
    We build the patch-to-patch adjacency matrix for segmentation based methods by neighborhood search:
    two segments $S_i, S_j$ are adjacent if and only if there exists $\mathbf{p}\in S_i, \mathbf{q}\in S_j$ such that $\mathbf{p}$ and $\mathbf{q}$ are within each other's 6-nearest neighbors.
    For our results, the binary patch-to-patch matrix is derived from the patch-to-curve matrix $\mathbf{FE}$, \ie $\mathbf{FF} = \left(\mathbf{FE}\times \mathbf{FE}^T \geq 1\right)$, where the inequality is checked component-wise.
\end{itemize}

\begin{figure*}[t]
    \centering
    \begin{overpic}[width=\linewidth]{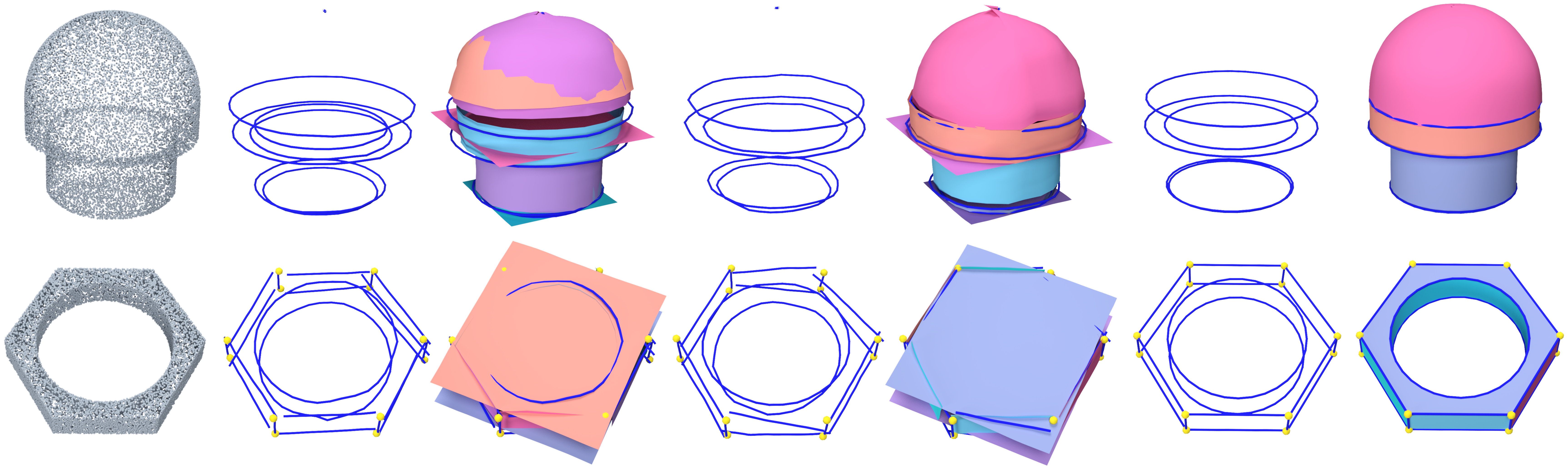}
        \put(3,0.3){\small Input points}
        \put(25,0.3){\small Baseline}
        \put(54,0.3){\small Ours-Net}
        \put(83,0.3){\small Ours-All}
        \put(0,27){\small (a)}
        \put(0,11){\small (b)}
    \end{overpic}
    \vspace{-6mm}
    \caption{\textbf{Ablation results}. We test the impact of the different components of our framework. For network predictions, we round the validness of elements by 0.5 to obtain the above results. Compared with a baseline detection network, \netname generation contains fewer redundancies and more accurate elements. Our chain complex extraction and geometric refinement further turn the network predictions into clean and complete B-Rep models. }
    \label{fig:ablation}
\end{figure*}
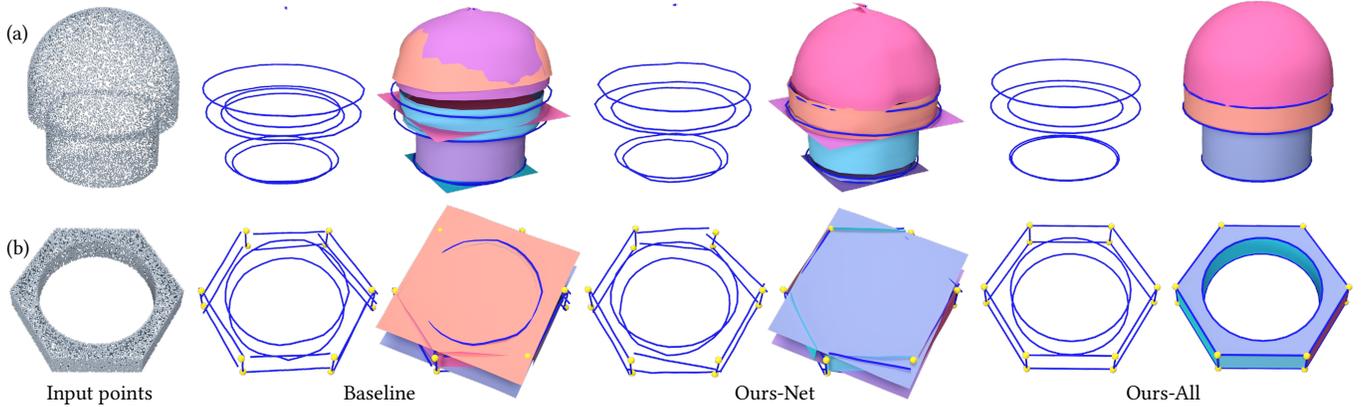

\begin{table*}[tb]
	\centering
	\caption{\textbf{Statistics of test results for ablation settings.} The baseline approach detects corners, curves and patches separately and suffers from low geometric and topological accuracies. Ours-Net improves the baseline by modeling topology explicitly. Ours-All further improves the network predictions with guaranteed topology consistency and highly improved geometric fitness (Fig.~\ref{fig:ablation}).}\vspace{-3mm}
	\begin{tabular}{c|c|ccc|ccc|ccc|ccc}
		\hline
		Config & Corner$\uparrow$ & \multicolumn{3}{c}{Curve$\uparrow$} & \multicolumn{3}{|c}{Patch$\uparrow$} & \multicolumn{3}{|c}{Topology error$\downarrow$} & \multicolumn{3}{|c}{Topo inconsistency$\downarrow$}\\
		 & F-score & F-score & Type acc & Open acc & F-score & Type acc & Open acc & FE & FV & EV &  |(\ref{eq:opt:cons:manifold})| & |(\ref{eq:opt:cons:curveend})| & |(\ref{eq:opt:cons:closed})| \\
		\hline
		Baseline    &77.0	&75.0	&76.7	&94.8	&77.1	&75.2	&93.4	&0.229	&0.213	&0.244	&4.51	&4.66	&2.47 \\
		Ours-Net    &80.9	&75.2	&76.9	&94.8	&78.8	&76.2	&93.6	&0.145	&0.131	&0.174	&0.873	&0.608	&0.154 \\
        Ours-Topo &81.8	&75.2	&76.7	&87.7	&81.3	&75.5	&93.4	&0.202	&0.171	&0.232	&0	    &0	    &0 \\
		Ours-All    &80.8	&74.7	&76.3	&87.8	&80.1	&74.2	&92.9	&0.201	&0.172	&0.230	&0		&0		&0 \\
		\hline
	\end{tabular}
	\label{tab:ablation}
\end{table*}

\subsection{Ablation Study}
\label{sec:ablation}

We do ablation study on major components of the whole pipeline, including network design and global optimization.
Starting from a baseline approach, we show the necessity and impact of different components by incorporating them incrementally.
\begin{enumerate}[leftmargin=15pt,itemsep=1pt]
    \item[(a)] \textbf{Baseline}. As a strong baseline, we use three separate transformer networks to detect the three groups elements with their validness, type classification and geometry embedding (removing the operation (\ref{eq:decode_step2}) effectively). 
    The topology adjacency can only be computed from geometric proximity of elements by rounding the matrices in (\ref{eq:geom_prox_topo}) by 0.5.
	\item[(b)] \textbf{Our network (Ours-Net)}.
	We use our \netname network design that enables the cross communication of three element groups and outputs categorical topology directly.
	The categorical predictions are simply rounded for obtaining element validness, type and topological adjacency.
	\item[(c)] \textbf{Our network with optimization}.
	We apply the neurally guided global optimization to the network predictions of (b), to obtain the topologically correct B-Rep chain complexes that have refined geometry, denoted as \textbf{Ours-All}.
	In addition, to discern the impacts of chain complex extraction and geometric refinement, we also report the intermediate results between these two steps, denoted as \textbf{Ours-Topo}.
\end{enumerate}
Visual and quantitative results are shown in Fig.~\ref{fig:ablation} and Table~\ref{tab:ablation}.

\paragraph{Explicit topology enables robust B-Rep reconstruction}

Topology reconstruction is critical for CAD reconstruction, not only because it is an essential component of the B-Rep structure and therefore a reconstruction target by itself, but also because through the modeling of topology the geometric reconstruction of B-Rep structure can be enhanced, during both network learning and geometric optimization.
This can be seen through the comparison of (a) the baseline prediction and (b) our network prediction.

The baseline network is a straightforward approach that uses three separate transformer decoders for the detection of corners, curves and patches respectively. 
On the other hand, our \netname additionally requires that the detected elements be consistent with each other through topology prediction and tri-path communication.
As a result, for all three element groups, our network detection accuracy is higher than the baseline (Table~\ref{tab:ablation}); especially for the corner detection, our F-score is higher by a large margin, as the corners (containing smooth junctions of multiple patches) are not prominent geometric features in the input point cloud, and their detection can be enhanced by the mutual constraints imposed by curves and patches that connect to them, as our network does.
Visually for the two examples shown in Fig.~\ref{fig:ablation}, the baseline network produces more redundant curves and patches than our network, which decreases its detection accuracy.

In terms of topology reconstruction, since the baseline network does not model or supervise topology learning directly, the topology relations can only be obtained from predicted geometry embeddings, which do not permit reconstruction of very accurate or consistent topology.
As shown in Table~\ref{tab:ablation}, the baseline results have larger topology errors from ground-truth and significantly larger errors in consistency than our results.

\paragraph{Global optimization promotes structure validness}

The naive rounding of our network predictions generally fails to comply with the B-Rep structure constraints (\ref{eq:opt:cons:manifold})-(\ref{eq:opt:cons:closed}), as shown by the topology inconsistency errors in Table~\ref{tab:ablation}.
The predicted geometry is also not tightly fitted to the input point cloud, and the different elements do not have conforming shapes, as shown in Fig.~\ref{fig:ablation}.
In comparison, by applying the second step of our framework, \ie solving the neurally guided combinatorial optimization and geometric fitting to the input point cloud and to constraints imposed by adjacent elements,
the results of our whole pipeline are complete and clean, as shown in Fig.~\ref{fig:ablation} and the zero topology inconsistency errors in Table~\ref{tab:ablation}.

We note that on average the global optimization changes detection accuracy from network prediction slightly in diverse ways, \eg the patch accuracy is improved while the curve accuracy is reduced, and changes the topology slightly farther away from ground-truth, which is expected as it enforces the topology constraints strictly to obtain a complete B-Rep structure and has to deviate more or less from the unconstrained network estimations.

Comparing the intermediate results of Ours-Topo with the final results of Ours-All, we find that the topology optimization indeed extracts consistent structures from network predictions, as shown in the topological errors of Table~\ref{tab:ablation}, while the geometric refinement step further introduces slight deviation of predicted elements so that they can fit together according to the solved topology, as evidenced by the slight decrease of F-scores from Ours-Topo to Ours-All.

\subsection{Comparison}

Since we take a holistic approach to the B-Rep structured CAD reconstruction, to our best knowledge we are not aware of any previous works that aim at the same output.
Nevertheless in this section, we compare our method with representative CAD reconstruction works that output geometric primitives where applicable, to discuss the relative strengths of different approaches.
We mainly compare with works that take the ``segmentation+primitive fitting'' paradigm for patch reconstruction \cite{Schnabel2007,yan2021hpnet,ParseNet2020ECCV,LiPrimitive2019CVPR}, as well as for detecting wireframes containing sharp corners and curves \cite{Wang2020pienet}.

\begin{figure*}
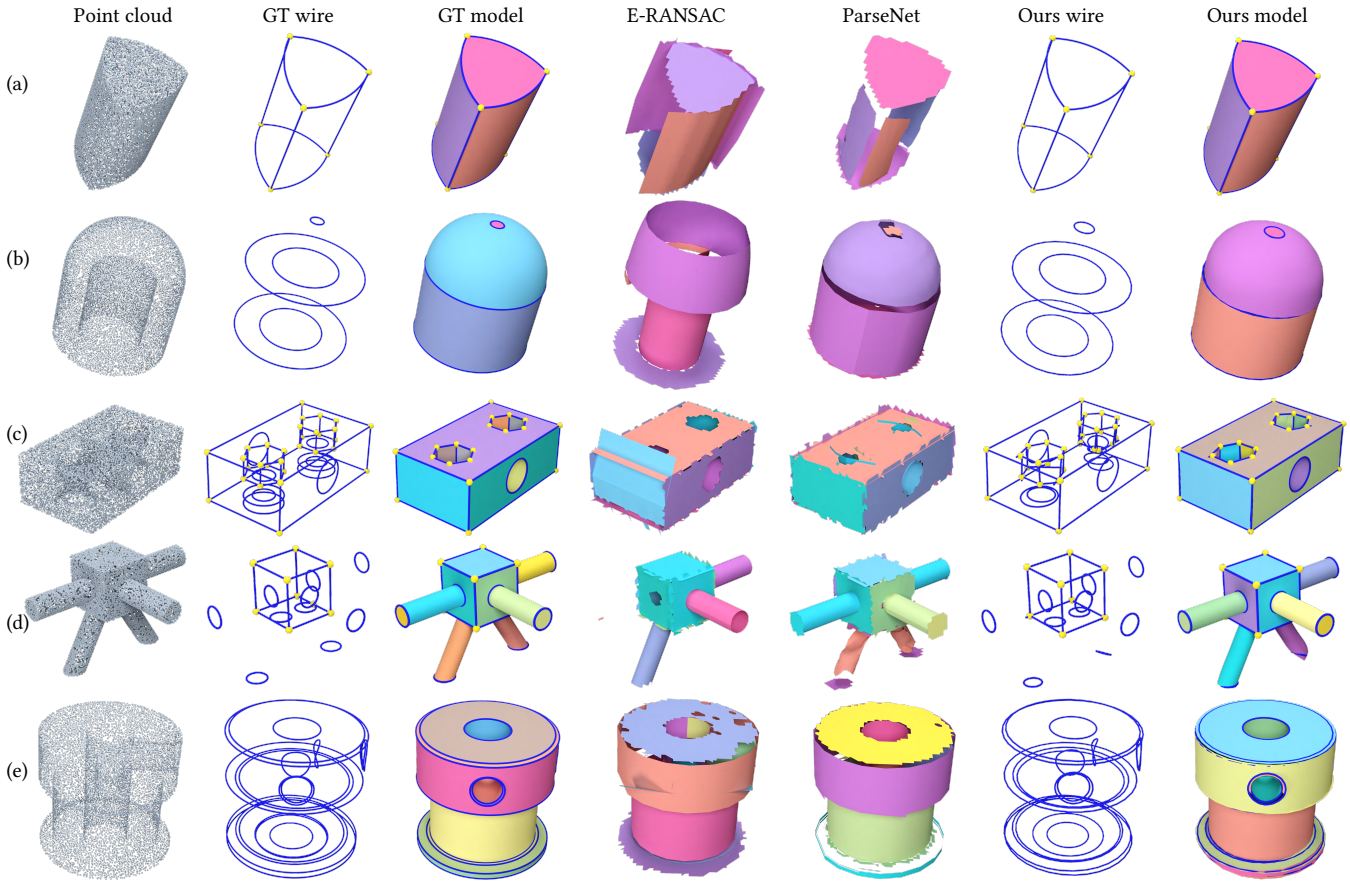

    \centering
    \vspace{4mm}
    \begin{overpic}[width=\linewidth]{image/comparison_revision}
	\put(5,64){\small Point cloud}
	\put(19,64){\small GT wire}
	\put(32,64){\small GT model}
	\put(46,64){\small E-RANSAC}
	\put(62,64){\small ParseNet}
	\put(75,64){\small Ours wire}
	\put(89,64){\small Ours model}
	\put(0,59){\small (a)}
	\put(0,46){\small (b)}
	\put(0,33){\small (c)}
	\put(0,19){\small (d)}
	\put(0,8){\small (e)}
	\end{overpic}
	\vspace{-7mm}
    \caption{\textbf{Comparison}. We compare our method with ``patch segmentation+fitting'' approaches including Efficient RANSAC and  ParseNet\protect\footnotemark. We note that while the patch fitting and trimming modules of comparing segmentation-based methods can be improved, our main difference from their results is the generation of three orders of elements that are consistently connected, which form more complete reconstructions that are closer to the structures of ground-truth.}
    \label{fig:comparison}
    \vspace{-1mm}
\end{figure*}

\footnotetext{HPNet \cite{yan2021hpnet} is not included in the visual comparison because the publicly shared code does not generate patches and the data does not contain patches either.}

\begin{table}[tb]
	\centering
	\caption{\textbf{Comparison with ``patch segmentation+fitting'' approaches}. The metrics evaluate patch fitting accuracy (residual), coverage and recall rate, as well as topology fidelity\protect\footnotemark.
		P-cov and P-to-P stand for p-coverage and patch-to-patch topology error, respectively.
	}
	\label{tab:comparison}
	\vspace{-3mm}
	\begin{tabular}{ccccc}
		\hline
		Method & Residual$\downarrow$ & Recall(\%)$\uparrow$ & P-cov(\%)$\uparrow$ & P-to-P$\downarrow$ \\
		\hline
		E-RANSAC          &0.022		&68.3   &83.4	&0.514 \\
		SPFN            &0.021		&-		&88.4	&- \\
		ParseNet        &0.011		&79.7	&93.0	&0.376 \\
		HPNet           &\textbf{0.009}		&81.8	&94.3	&0.343 \\
		Ours            &0.019		&\textbf{87.9}	&\textbf{95.6}	&\textbf{0.191} \\
		\hline
	\end{tabular}
\vspace{-1mm}
\end{table}

\footnotetext{The recall and patch-to-patch error for SPFN \cite{LiPrimitive2019CVPR} are not reported as the evaluation is based on an extension of SPFN done by ParseNet \cite{ParseNet2020ECCV} that incorporates splines and is not released with sufficient details. For the same reason SPFN results are not visualized in Fig.~\ref{fig:comparison}.}

\paragraph{With ``patch segmentation+fitting'' approaches}

In this category of methods, except for Efficient RANSAC \cite{Schnabel2007} which is a search algorithm, the other three approaches are all learning-based methods trained on the same split of ABC dataset as ours.
In addition, all results are obtained on the same test set containing input point clouds with normal vectors.

As reported in Table~\ref{tab:comparison}, on metrics of surface patch accuracy and coverage, our results are comparable to or surpass state-of-the-art patch segmentation based methods. 
For example, the residual error of our results is between that of ParseNet \cite{ParseNet2020ECCV} and SPFN \cite{LiPrimitive2019CVPR}, but our p-coverage is higher than the other methods.
Visually as shown in Fig.~\ref{fig:comparison}, our results always have complete structures that contribute to the higher coverage.

Note that the residual computation discards unmatched GT segments (patches) from consideration.
When comparing our results with the other methods, we find that GT patches generally have much better coverage and mostly have matched prediction patches, which makes our per-patch residual error more prone to larger values.
To reveal this fact, in Table~\ref{tab:comparison} we also report the recall rates of GT patches that are matched to predictions, which shows a much larger recall rate of our result than others.
Additionally, if we apply a filtering of patches with outlier residual errors (\eg residual $> 0.05$) for our result, the remaining average residual error is reduced to 0.0088 (slightly smaller than the lowest error by HPNet \cite{yan2021hpnet}), while the recall rate is 81.1\% and still comparable to the best of previous results.

Besides patch accuracy, we also evaluate the patch-to-patch topology error which measures topological fidelity to ground truth.
As shown in Table~\ref{tab:comparison}, the segmentation based approaches generally have larger errors from GT topology than our result.
The difference can be understood intuitively by inspection of concrete examples.
Visually as shown in Fig.~\ref{fig:comparison}, the segmented patches frequently do not cover all regions of the input point cloud nor conform with each other, therefore not forming structurally valid B-Rep models.
In contrast, our approach generally reconstructs structurally consistent and more complete B-Rep models.

\paragraph{With PIENET \cite{Wang2020pienet}}
PIENET targets on \textit{sharp features} only by taking a two-stage process that detects sharp curve and corner points first and then generates parametric curves that connect up the first stage points.
Since PIENET is also trained on ABC data, we directly evaluate its trained network on our test point clouds.

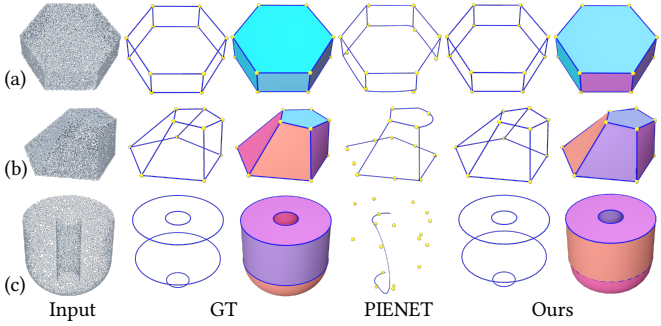
\begin{figure}
    \centering
    \begin{overpic}[width=\linewidth]{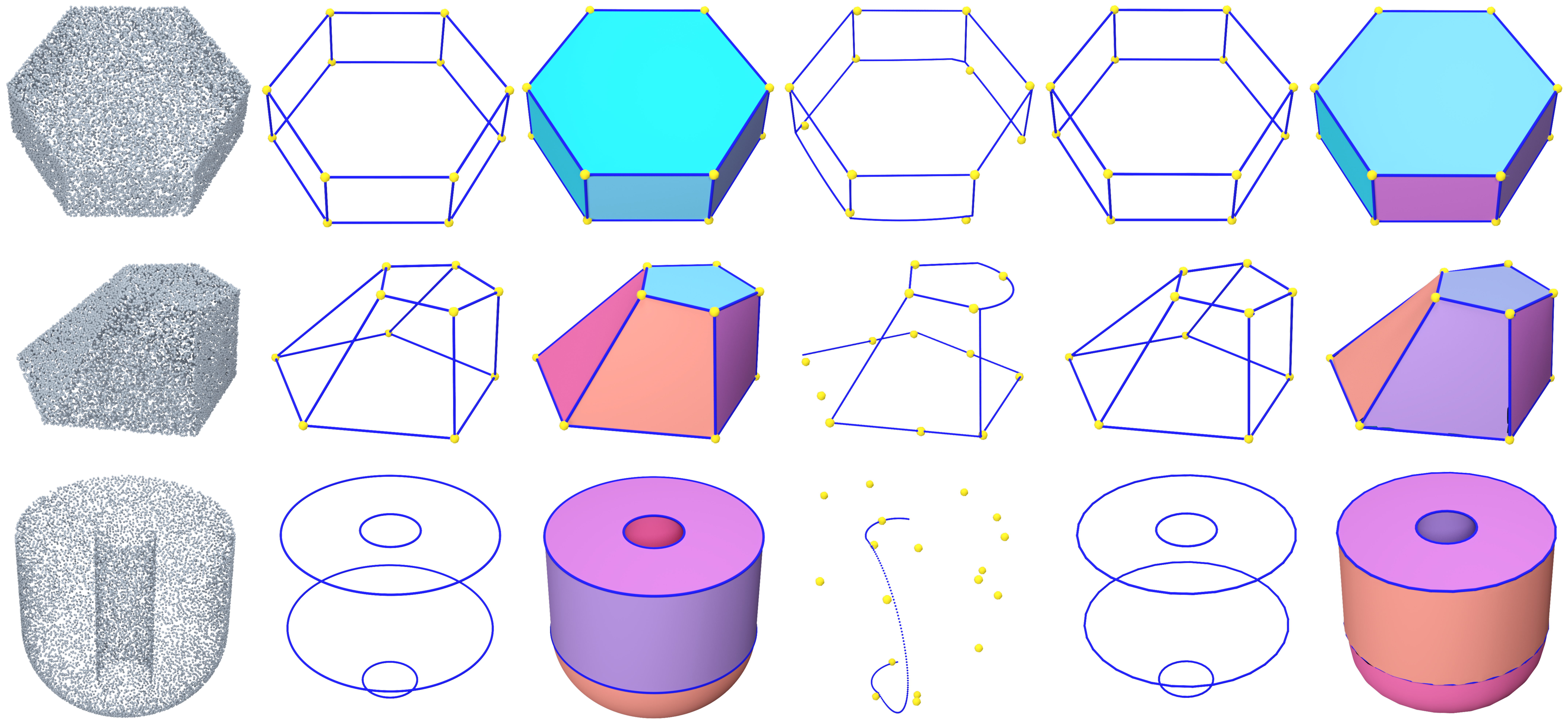}
        \put(5,-3){\small Input}
        \put(30,-3){\small GT}
        \put(54,-3){\small PIENET}
        \put(80,-3){\small Ours}
        
        \put(-2,33){\small (a)}
        \put(-2,19){\small (b)}
        \put(-2,1){\small (c)}
    \end{overpic}
    \caption{\textbf{PIENET comparison.} We compare our method with PIENET that detects sharp corners and curves. 
    For (a)(b), our complex structure ensures that the curves and corners connect into valid patch boundaries and therefore more complete.
    For (c), we detect smooth curves in addition to sharp ones, as smooth curves delineate primitive patches. }
    \label{fig:comparison_PIENET}
\end{figure}

Qualitative cases are shown in Fig.~\ref{fig:comparison_PIENET}, where we see that PIENET results may not form complete curve networks of sharp features, while our result curves and corners have to connect into closed patch boundaries and are therefore more likely to be complete.
In addition, our results contain both smooth curves and sharp features, as smooth curves (corners) are needed to delineate and connect the primitive patches.
The comparison demonstrates that our B-Rep chain complex structure promotes more complete and regular reconstructions.
Due to the significant difference of reconstruction targets, we deem a qualitative comparison for the two methods uninformative and do not include it in the paper.

\subsection{Stress Tests}
We further evaluate the performance of \name under more challenging cases with data corruptions or generalization to unseen data.
For corrupted data test, we generate noisy and partial point clouds respectively, and retrain our network on these two datasets for evaluation.
For reference, we also train the state-of-the-art ParseNet \cite{ParseNet2020ECCV} model on these datasets, although we note that these tasks are not intended to be addressed by such a segmentation based approach, as a corrupted point cloud poses challenge for dense segmentation.
For generalization tests, we apply trained networks to real scans and daily objects from ShapeNet~\cite{shapenet2015} that are categorically different from the mechanical parts used for training. 

\paragraph{Noisy data}

To simulate real noisy data, we apply normal direction offset for each point with offset distance independently sampled from two normal distributions of zero mean and standard deviations $\sigma_{1} = 0.02$ and $\sigma_{2} = 0.05$ respectively, which generates two datasets for training and testing.
In addition, we omit the normal vectors of input points for both training and testing of all approaches.

As reported in Table~\ref{tab:comparison_noise_partial}(a)(b), given the noisy input, both approaches have degraded patch accuracy.. 
However, the impact on our approach is much smaller compared with that on ParseNet throughout all metrics.
Visually as shown in Fig.~\ref{fig:noise_partial_data}(a)(b), we find that for such noisy data ParseNet has difficulty to obtain reliable and complete segmentations and patches, which is understandable as a segmentation-based approach highly relies on the regularity of local point distributions for segmentation boundary detection and patch fitting.
In contrast, even though the delicate input features may be too corrupted to tell (\eg Fig.~\ref{fig:noise_partial_data}(b) the inner polygonal loop), thanks to our holistic modeling of the chain complex B-Rep structure, our results can leverage the mutual constraints among elements and still define complete reconstructions with strong regularity.

\paragraph{Partial data}
To simulate partial data, we virtually scan each object in our dataset with $3{\sim}4$ view points randomly distributed over the 8 corners of the object bounding box.
The partial data frequently misses corners and inner structures of an object (see Fig.~\ref{fig:noise_partial_data}(c)(d)).

As reported in Table~\ref{tab:comparison_noise_partial}(c), both approaches are again impacted by the partiality of the input points.
Notably, the residuals of both approaches do not suffer that much, but the coverage of ParseNet results is significantly degraded than normal input (Table~\ref{tab:comparison}).
In comparison, due to our modeling of complete structures, the coverage of our results does not suffer obviously than the normal input. 
Once again as in the normal input case, our residual is likely to be biased by the much higher coverage.
Visually as shown in Fig.~\ref{fig:noise_partial_data}(c)(d), we find that given partial data the patches of ParseNet results are more difficult to fit well and generally are misaligned. 
In comparison, our results still maintain a complete and regularized structure even though it may be different from the ground truth.

\begin{figure}
    \centering
    \begin{overpic}[width=\linewidth]{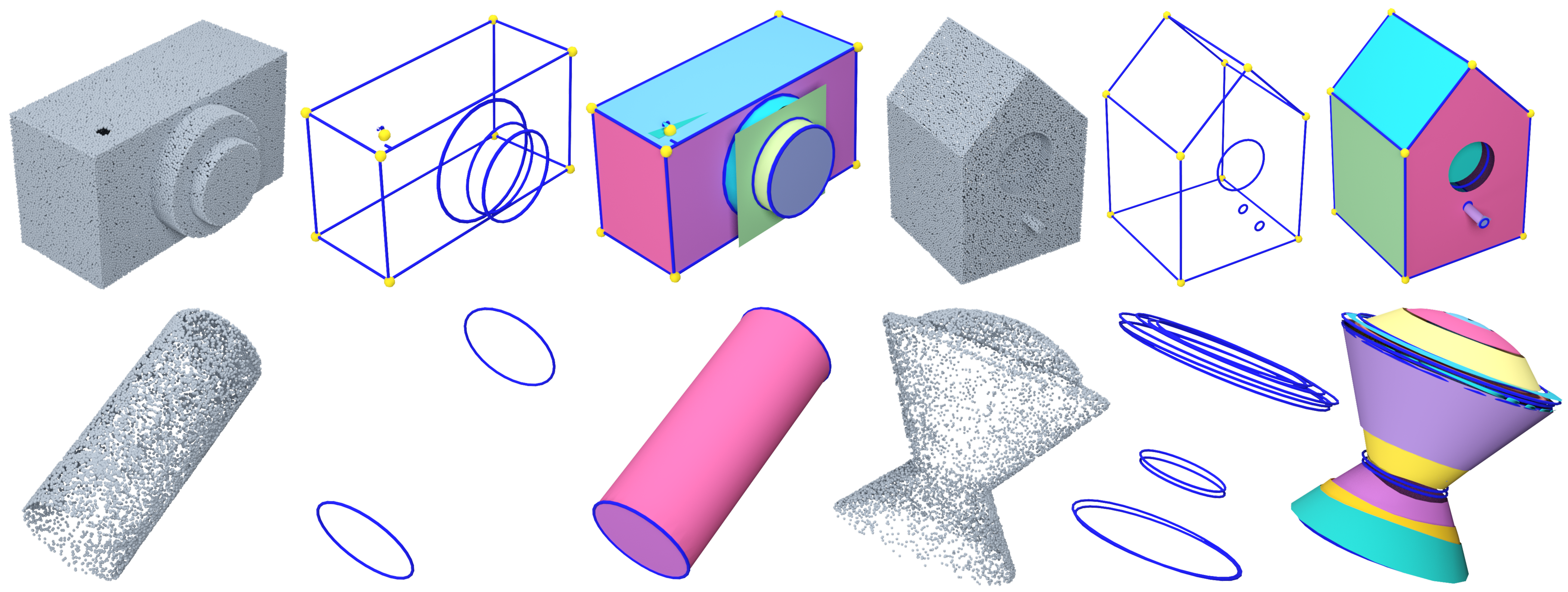}
    \end{overpic}
    \vspace{-5mm}
    \caption{\textbf{Generalization tests} on ShapeNet models (top row) and real scans from AIM@SHAPE-VISIONAIR repository (bottom row). The real scans lack bottom parts and have highly nonuniform sample points.}
    \label{fig:realscan_shapenet}
\end{figure}

\paragraph{Out-of-distribution generalization tests}
ShapeNet models are daily objects with different characteristics from mechanical parts, but we apply the network trained with clean ABC data on sampled points of ShapeNet models for generalization test.
Real scans generally contain missing regions, so we apply the network trained with synthetic partial data on them.
Examples are shown in Fig.~\ref{fig:realscan_shapenet}, where we find that the synthetically trained networks can still process these out-of-distribution shapes. 
Meanwhile, the fragmented predictions of a real-scanned doorknob show that our training is overly focused on subtle geometric variations of mechanical parts; such train/test discrepancy may be mitigated by stronger data augmentation and weakly supervised learning on diverse shape collections (\cf Sec.~\ref{subsec:loss_functions}).

\begin{table}[tb]
	\centering
	\caption{\textbf{Comparison in quantitative metrics under stress tests}. We compare our results with ParseNet on noisy or partial input point clouds. }
	\label{tab:comparison_noise_partial}
	\vspace{-3mm}
	{\small (a) Noisy data with $\sigma_1=0.02$} 
	\begin{tabular}{ccccc}
		\hline
		Method & Residual$\downarrow$ & Recall(\%)$\uparrow$ & P-cov(\%)$\uparrow$ & P-to-P$\downarrow$ \\
		\hline
		ParseNet   &0.041	&52.4	&53.3	&0.673\\
		Ours       &\textbf{0.029}	&\textbf{84.2}	&\textbf{83.2}	&\textbf{0.255}\\
		\hline
	\end{tabular}\\
	
	\vspace{1mm}
	{\small (b) Noisy data with $\sigma_2=0.05$} 
	\begin{tabular}{ccccc}
		\hline
		Method & Residual$\downarrow$ & Recall(\%)$\uparrow$ & P-cov(\%)$\uparrow$ & P-to-P$\downarrow$ \\
		\hline
		ParseNet   &0.058	&54	    &44.8	&0.686\\
		Ours       &\textbf{0.038}	&\textbf{77.4}	&\textbf{69.9}	&\textbf{0.302}\\
		\hline
	\end{tabular}\\
	
	\vspace{1mm}
	{\small (c) Partial data}
	\begin{tabular}{ccccc}
		\hline
		Method & Residual$\downarrow$ & Recall(\%)$\uparrow$ & P-cov(\%)$\uparrow$ & P-to-P$\downarrow$ \\
		\hline
		ParseNet & \textbf{0.018} & 78.3 & 76.9 & 0.410 \\
		Ours  & 0.025 & \textbf{85.2} & \textbf{93.1} & \textbf{0.223} \\
		\hline
	\end{tabular}
\end{table}

\begin{figure*}
	\centering
	\begin{overpic}[width=0.9\linewidth]{image/partial_noise_compact}
		\put(-1.5,50){\small (a)} 
		\put(-1.5,37){\small (b)} 
		\put(-1.5,23){\small (c)} 
		\put(-1.5,9){\small (d)} 
		\put(3,-2){\small Input points} 
		\put(22,-2){\small GT wire} 
		\put(38,-2){\small GT model} 
		\put(57,-2){\small ParseNet} 
		\put(75,-2){\small Ours wire} 
		\put(89,-2){\small Ours model} 
	\end{overpic}
	\caption{\textbf{Stress tests with noisy and partial data}. (a)(b) are noisy inputs corrupted with $\sigma_1=0.02$, (c)(d) are partial inputs.
		By generating the different orders of elements simultaneously that are mutually constrained, our reconstruction shows more robustness and the results are more complete than those of ParseNet which is a segmentation based approach and likely to be more sensitive to local point sampling quality. }
	\label{fig:noise_partial_data}
\end{figure*}

\subsection{Validness Assessment Results}
\label{sec:validness_results}

\begin{figure}
    \centering
    \begin{overpic}[width=0.7\linewidth]{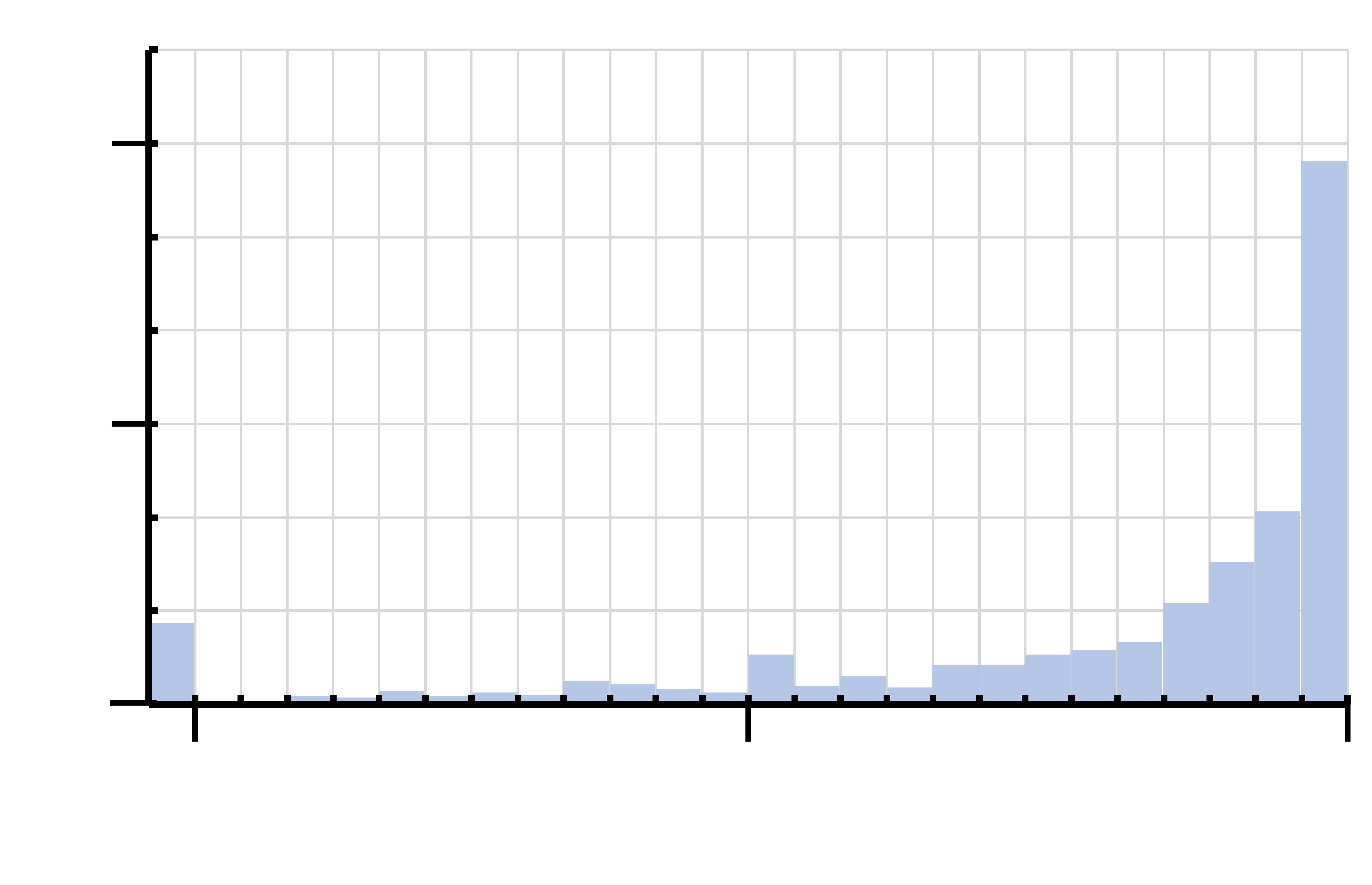}
        \put(37,3){\small Sample validness ratio}
        \put(-9,59){\small \#Samples}
        \put(10,7){\small $\leq 50$}
        \put(52,7){\small $75$}
        \put(95,7){\small $100$ \ \%}
        \put(5,13){\small 0}
        \put(1,33){\small 600}
        \put(-1,53){\small 1200}
    \end{overpic}
    \vspace{-3mm}
    \caption{\textbf{Validness distribution} for all 3k test samples. 
    Using a threshold of $0.03$, we find most of the test samples have high ratios of valid topology connections fulfilled by geometric embeddings. See Fig.~\ref{fig:validness_example} for a concrete analysis of partially valid sample.}
    \label{fig:validness_distribution}
\end{figure}

\begin{figure}
	\centering
	\begin{overpic}[width=\linewidth]{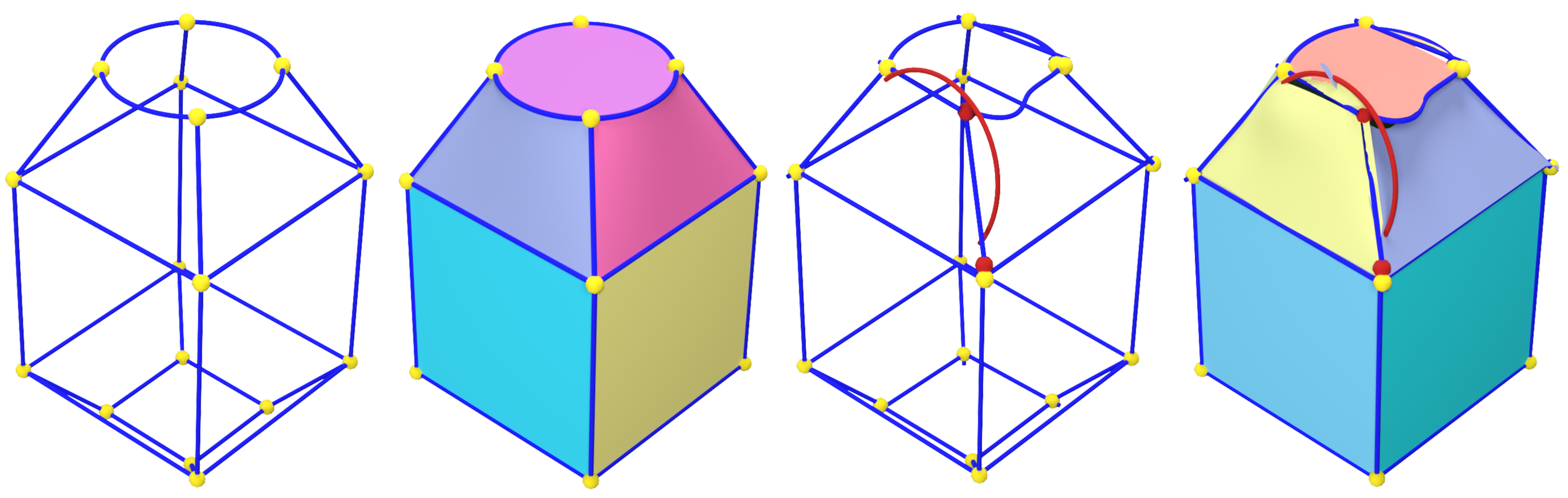}
		\put(17,-2){\small (a) GT model}
		\put(67,-2){\small (b) Our result}
	\end{overpic}
	\caption{\textbf{Validness analysis}. This example has 97\% validness. The mismatch comes from the curves and corners highlighted in red, which are connected according to topology but cannot be realized geometrically as the curve is predicted to be a circular arc. The curve and one of its corner are actually redundant compared with ground-truth. The lower part of the model is valid and the same as GT.}
	\label{fig:validness_example}
\end{figure}

As discussed in Sec.~\ref{sec:validity_checking}, since we always have a valid topology connection, we can assess the result quality by checking how well the geometry embedding fulfills the prescribed topology.
When we apply a distance threshold of 0.03, the distribution of validness ratio for all test samples is shown in Fig.~\ref{fig:validness_distribution}.
We can see that a large portion of our results concentrate on the high validness region.
Furthermore, as shown in Fig.~\ref{fig:validness_example}, the violation of validness usually pinpoints to a specific structural issue of the reconstructed model.
The causes for the issues can be problematic network predictions, or being stuck to a local optimal solution for either the chain complex extraction or the geometric refinement.
To find repairing solutions (possibly with a minimum amount of user interaction) under the guidance of validity assessment is an important future work.

\subsection{Limitations and Future Work}
\label{sec:limitation}

To our best knowledge, this work represents a first effort on reconstructing holistic CAD B-Rep structures from point clouds directly.
While we have showed that such an ambitious target and comprehensive formulation can be beneficial for more complete and structured reconstruction, each step of the specific implementation we have taken can almost surely be further improved.

The network detection may miss important geometric features especially when they are small.
To this end, we would like to explore how to combine the segmentation based approach with our generative approach so that geometrically interesting features can be reliably incorporated into the complex generation procedure.
Possible solutions could be to propose elements based on a first pass of geometric segmentation; for 2D image detection adaptive proposal generation is shown to improve training convergence and result quality \cite{meng2021CondDETR}.

Both complex extraction and geometric refinement steps can be enhanced.
For example, the objective function of complex extraction (Sec.~\ref{sec:topo_opt}) can use more hints, including the correspondence to input points for better geometric data integration and the fine-tuning of term weights based on prediction accuracy on validation sets.
On the other hand, as noted in Sec.~\ref{sec:geom_opt}, more constraints (CAD operations) can be inferred from the typed primitives and their mutual topology, as shown in \cite{ZoneGraph2021,lambourne2021brepnet}.

\section{Conclusion}

We have presented \name, a framework that consists of a holistic representation of CAD B-Rep models as chain complexes and a learning plus optimization based approach for generating such structured representation from input point clouds.
The chain complex representation naturally introduces structural constraints to enhance model validness.
The learning based approach uses a transformer-based detection network with three paths for all element groups to generate the geometric primitives, \ie corners, curves and patches, as well as their mutual topological relationships simultaneously;
such a design is shown to improve detection and topology reconstruction than the baseline approach without topology generation.
The neurally guided optimization further finalizes the predicted complex by solving tractable combinatorial and geometric fitting problems with structure validness constraints as induced by the chain complex representation. 
Through extensive tests on large dataset, we demonstrate that such a structure-rich reconstruction task enables more complete and regular CAD B-Rep models being recovered.

\begin{acks}
We thank Yuqi Yang and Yuezhi Yang for help with ParseNet and RANSAC comparison. 
Real scans of Fig.~\ref{fig:realscan_shapenet} are provided courtesy of INRIA by the AIM@SHAPE-VISIONAIR Shape Repository.
\end{acks}

\bibliographystyle{ACM-Reference-Format}
\bibliography{src/complexgen}

\appendix

\section{Network structure details}
\label{appn:network_details}

\begin{figure}
    \centering
    \includegraphics[width=0.25\linewidth,angle=90]{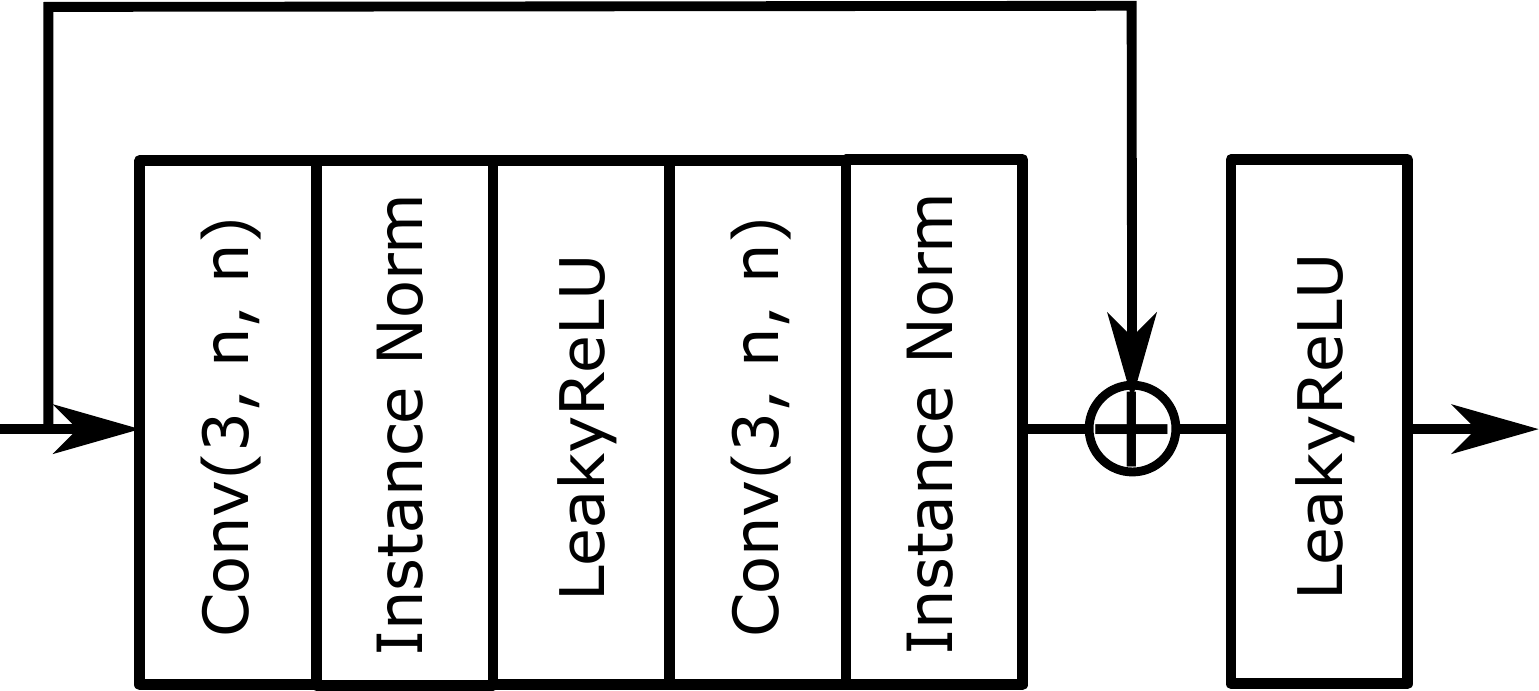}
    \vspace{-3mm}
    \caption{Residual block of the SparseCNN encoder. $n$ is the channel size.}
    \label{fig:encoder_resblock}
    \vspace{-5mm}
\end{figure}

The encoder network is a sparse CNN \cite{choy20194d} with residual connections. 
It contains 22 layers of convolution, pooling and instance normalization layers, organized as follows:
{\small
\abovedisplayskip=1pt
\abovedisplayshortskip=1pt
\belowdisplayskip=1pt
\belowdisplayshortskip=1pt
\begin{align}
    &\text{Conv(1, 7, 64)} -\\ &3\times\text{Block(64)} - \text{Conv(1, 64, 128)} - \text{Pooling} -\\
    &3\times\text{Block(128)} - \text{Conv(1, 128, 256)} - \text{Pooling} -\\
    &3\times\text{Block(256)} - \text{Conv(1, 256, 384)} - \text{Pooling} -\text{Block(384)},
\end{align}
}%
where Conv(\textit{kernel size, in channels, out channels}) denotes a 3D sparse convolution with specified kernel size, input channel size and output channel size, Pooling is max pooling by a factor of 2, and Block(\textit{channels}) is a residual block with structure shown in Fig.~\ref{fig:encoder_resblock}.

The tri-path transformer decoder network for generating the elements of different orders has $l=6$ layers.
Each self-attention or cross attention operation has 8 attention heads that split the latent vector equally.
No dropout is used as we find it causes instability for our regression task.

The geometry embedding modules are hypernets that are modulated by latent element vectors and map from the unit parameter domains to spatial points (Fig.~\ref{fig:geom_embed}).
In particular, the latent vector is first projected to 128 dimension, and then generates an MLP with 3 layers and feature map sizes [$x$, 64, 64, 3], where $x$ is the parameter dimension (1 for curves, 2 for patches).
LeakyReLU activation is used in the hidden layers, as the curves and patches do not have high frequency details to generate.

The projection heads for validness and other classification tasks are three-layer MLPs with fixed intermediate dimension $d=384$.
The projection heads for topology generation are FC layers that project the latent vector to 256 dimension before computing dot product correlation.

\section{ILP formulation of the structure extraction problem}
\label{appn:struct_extraction_detail}

We use the following trick to convert quadratic terms of binary variables into variables with linear constraints only.
For two binary variables $x,y$, their product $xy\in \{0,1\}$ can be represented by another binary variable $z$ with the following constraints:
\begin{equation}
	z \leq x,\quad z \leq y,\quad z \geq x + y - 1. \nonumber
\end{equation}
The problem (\ref{eq:chain_extraction_problem}) of Sec.~\ref{sec:topo_opt} is therefore reformulated as 
\begin{align}
	\textbf{max} &\quad w F_{topo} + (1-w) F_{geom} \\
	\textbf{s.t.} &\quad \textstyle{\sum}_{i}{\mathbf{FE}[i,j]} = 2\mathbf{E}[j],  \nonumber\\
	&\quad \begin{cases}
		\mathbf{Y}[i] \leq \mathbf{E}[i],\\
		\mathbf{Y}[i] \leq \mathbf{O}[i],\\
		\mathbf{Y}[i] \geq \mathbf{E}[i] + \mathbf{O}[i] - 1,\\
		\sum_{j}{\mathbf{EV}[i,j]} = 2\mathbf{Y}[i],
	\end{cases} \nonumber\\
	&\quad \begin{cases}
		\mathbf{Z}[i,j,k] \leq \mathbf{FE}[i,j],\\
		\mathbf{Z}[i,j,k] \leq \mathbf{EV}[j,k],\\ 
		\mathbf{Z}[i,j,k] \geq \mathbf{FE}[i,j] + \mathbf{EV}[j,k] - 1,\\
		\sum_{j}\mathbf{Z}[i,j,k] = \mathbf{FV}[i,k],
	\end{cases} \nonumber\\
	&\quad \begin{cases} 
		\mathbf{FE}[i,j] \leq \mathbf{F}[i] \left(\leq \sum_j{\mathbf{FE}[i,j]}\right), \\
		\mathbf{EV}[i,j] \leq \mathbf{V}[j] \leq \sum_{k}{\mathbf{EV}[k,j]}, 
	\end{cases} \nonumber \\
	&\quad \mathbf{F} \in \mathbb{B}^{N_f}, \mathbf{E},\mathbf{O} \in \mathbb{B}^{N_e}, \mathbf{V} \in \mathbb{B}^{N_v}, \nonumber\\
	&\quad \mathbf{FE} \in \mathbb{B}^{N_f\times N_e}, \mathbf{EV} \in \mathbb{B}^{N_e\times N_v}, \mathbf{FV} \in \mathbb{B}^{N_f\times N_v}, \nonumber\\ 
	&\quad \mathbf{Y} \in \mathbb{B}^{N_e}, \mathbf{Z} \in \mathbb{B}^{N_f\times N_e \times N_v} \nonumber
\end{align}
where we have introduced the binary variables $\mathbf{Y}, \mathbf{Z}$ and corresponding linear constraints to replace the quadratic constraints (\ref{eq:opt:cons:curveend}) and (\ref{eq:opt:cons:closed}). 
Note that the element variables should be truncated by their predicted validness probabilities (Sec.~\ref{sec:topo_opt}) but we abuse notations slightly and still denote the remaining numbers as $N_v, N_e, N_f$.

\end{document}